\documentclass[preprint,12pt,authoryear]{elsarticle}

\usepackage{amssymb}

\journal{Nuclear Physics B}
\biboptions{numbers,sort&compress}

\label{key}
\usepackage{listings}
\usepackage{graphicx}
\usepackage{stfloats}
\usepackage{xcolor}
\usepackage{colortbl,booktabs}
\usepackage{bigstrut,bigdelim,multirow}
\usepackage{algorithm}
\usepackage{algorithmic}
\usepackage{setspace}
\usepackage{etoolbox}
\usepackage{amsmath}
\usepackage{array}
\usepackage{makecell}
\usepackage{ragged2e}
\usepackage{epstopdf}
\usepackage{booktabs}
\usepackage{array, caption, threeparttable}
\usepackage[font=small,labelfont=bf,labelsep=none]{caption}

\usepackage{ragged2e}
\usepackage{amssymb}
\usepackage{pifont}
\usepackage{color}
\usepackage{wrapfig}
\usepackage{textcomp}
\usepackage{natbib}
\definecolor{lightgray}{gray}{0.93}
\definecolor{lightblue}{RGB}{220,235,250}
\definecolor{lightgray}{gray}{0.93}
\setcitestyle{numbers,square}
\usepackage[none]{hyphenat}

\def\tsc#1{\csdef{#1}{\textsc{\lowercase{#1}}}}
\tsc{WGM}
\tsc{QE}
\tsc{EP}
\tsc{PMS}
\tsc{BEC}
\tsc{DE}
\begin{document}
\begin{sloppypar}

\begin{frontmatter}

\title {\textbf{SGDFuse: SAM-Guided Diffusion Model for High-Fidelity Infrared and Visible Image Fusion}}

%\tnotetext[1]{This research was supported by the National Natural Science Foun dation of China (61772319, 62002200, 12001327, 62176140), Shandong Natural Science Foundation of China (ZR2020QF012 and ZR2021MF068).}

\author[1]{Xiaoyang Zhang}
%\cormark[1]
%\fnmark[1]
\ead{2023410090@sdtbu.edu.cn}
%\ead[url]{www.cvr.cc, cvr@sayahna.org}
%\credit{Conceptualization of this study, Methodology, Software}

\author[1,2]{Jinjiang Li}
%\cormark[1]
%\fnmark[1]
\ead{lijinjiang@sdtbu.edu.cn}
\author[1]{Guodong Fan}
%\cormark[1]
%\fnmark[1]
\ead{fanguodong@sdtbu.edu.cn}
\author[3]{Yakun Ju}
\ead{yj174@leicester.ac.uk}

\author[4]{Linwei Fan}
%\fnmark[2]
\ead{fanlinwei@sdufe.edu.cn}
%\fnmark[2]

\author[3]{Jun Liu}
\ead{j.liu81@lancaster.ac.uk}

\author[5]{Alex C. Kot}
\ead{eackot@ntu.edu.sg}

%\credit{Data curation, Writing - Original draft preparation}
%\address[1]{School of Information and electronic engineering, Shandong Technology and Business University, Institute of Network Technology (ICT), Yantai 264005, China}
\address[1]{School of Computer Science and Technology, Shandong Technology and Business University, Yantai, China.}
\address[2]{College of Computer and Data Science, Fuzhou University, Fuzhou, China.}
\address[3]{School of Computing and Mathematical Sciences, University of Leicester, Leicester, United Kingdom.}
%\address[3]{School of Software, Shandong University, Jinan 250101, China}
\address[4]{School of Computing and Artificial Intelligence, Shandong University of Finance and Economics, Jinan, China.}
%\address[4]{School of Computer Science and Informatics, Cardiff University, Cardiff, United Kingdom.}
\address[5]{Director of the Rapid-Rich Object Search Laboratory and the NTU-PKU Joint Research Institute, Nanyang Technological University, Singapore.}

% 通讯作者注释
\cortext[1]{Corresponding author: Guodong Fan (fanguodong@sdtbu.edu.cn).}
% \cortext[cor2]{Principal corresponding author}

\begin{abstract}
Infrared and visible image fusion (IVIF) is essential for integrating thermal saliency with textural details to support downstream perception. However, most existing approaches suffer from "semantic blindness," leading to the erroneous suppression of thermal targets and the introduction of visual artifacts. To address this, we propose SAM-Guided Diffusion Fusion Network (SGDFuse), a novel Semantic-Guided Generation (SGG) framework that reframes IVIF as a semantically-steered generative task rather than simplistic pixel mapping. Our method uniquely couples high-level semantic priors from the Segment Anything Model (SAM) with the high-fidelity generative power of a conditional diffusion model. We employ a deliberate two-stage strategy to decouple multimodal alignment from iterative refinement: Stage I establishes a robust structural foundation via preliminary fusion, while Stage II utilizes dual-modality semantic masks as spatial anchors to guide the diffusion process toward a semantically coherent, high-fidelity reconstruction. Comprehensive experiments demonstrate that SGDFuse not only delivers state-of-the-art image quality but also enhances downstream task performance, confirming its effectiveness as a new Methodological Framework for semantically aware image fusion. The code is available at https://github.com/boshizhang123/SGDFuse.
\end{abstract}

\begin{keyword}
Image Fusion\sep Diffusion Model\sep Segment Anything Model \sep Multimodal Information\sep Image Enhancement
\end{keyword}

\end{frontmatter}

%% \linenumbers

%% main text
\section{Introduction}
Infrared and visible image fusion is an important research direction in the field of computer vision \cite{1}. This technology aims to efficiently combine the thermal radiation information of infrared images with the rich texture details of visible light images, to generate an image that is more comprehensive in information and higher in perceptual clarity \cite{2}. Because infrared images can effectively overcome adverse conditions such as smoke and low light, while visible light images possess the advantages of high resolution and high contrast, the features of the two are highly complementary. Consequently, this technology has become a cornerstone in diverse fields. In the realm of smart healthcare, multimodal image fusion provides critical support for precise clinical diagnosis. For instance, a medical image fusion framework utilizing co-occurrence filters and local extrema in the NSST domain has been developed to enhance biological detail \cite{m1}. Other advanced methods include end-to-end content-aware generative adversarial networks \cite{m2} and clustering-based \cite{m3}, \cite{m4} approaches for robust medical synthesis in smart healthcare systems. Furthermore, frameworks like TSJNet \cite{m5} have explored target and semantic awareness to refine fusion quality. Beyond medical analysis, this technology also significantly improves environmental perception and target recognition in key scenarios such as autonomous driving \cite{3} and military reconnaissance, effectively enhancing the performance of downstream vision tasks \cite{5}. Therefore, how to effectively achieve fusion to balance the final visual quality and task requirements, is an urgent challenge currently being faced.

In pursuit of this objective, a plethora of techniques have been proposed over the past few decades \cite{6}. Conventional fusion methods are largely predicated on handcrafted feature extraction and fusion rules—such as multi-scale transforms \cite{7}, and subspace reconstruction—which offer a degree of interpretability and robustness. However, these approaches face inherent limitations in modeling the high-order semantic relationships and non-linear complementarities between modalities, thereby restricting their efficacy in complex scenes \cite{8}. With the advent and rapid proliferation of Deep Learning, the research landscape for IVIF has undergone a paradigm shift. This has yielded substantial advancements in feature extraction, fusion, and reconstruction, rapidly establishing the field as a prominent area of research. For instance, Li et al. introduced DenseFuse \cite{9}, which leverages an encoder-decoder architecture to extract multi-level features; PIAFusion \cite{10}, proposed by Tang et al., employs a progressive fusion strategy to handle extreme variations in illumination; Text-DiFuse \cite{11} achieves controllable fusion and enhanced robustness against degradation by integrating text modulation with diffusion-based generation.
\begin{figure}[!b]
\centering
\includegraphics[scale=0.8]{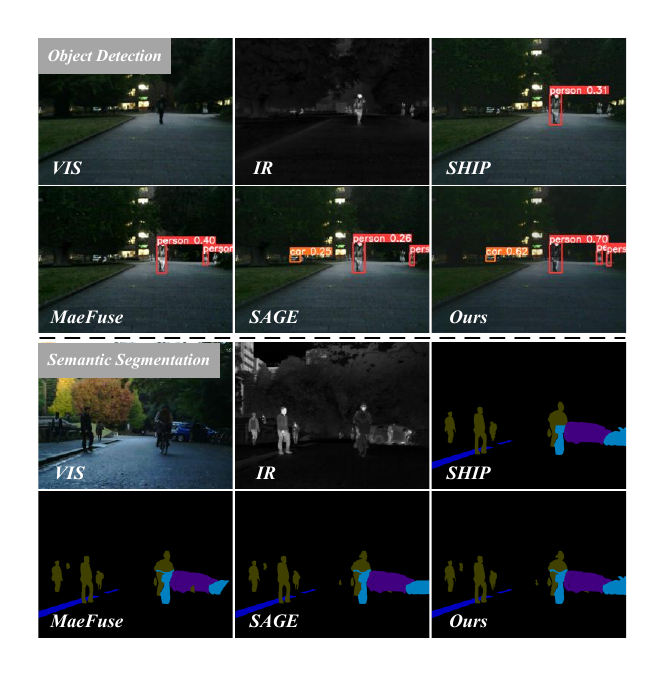}
\caption{Comparison of different fusion methods on object detection and semantic segmentation tasks. Traditional methods perform well in terms of visual quality but are limited in downstream tasks, while the proposed method achieves better task performance while maintaining image quality.}
\label{figure}
\end{figure}
\begin{figure}[htbp]
\centering
\includegraphics[scale=0.78]{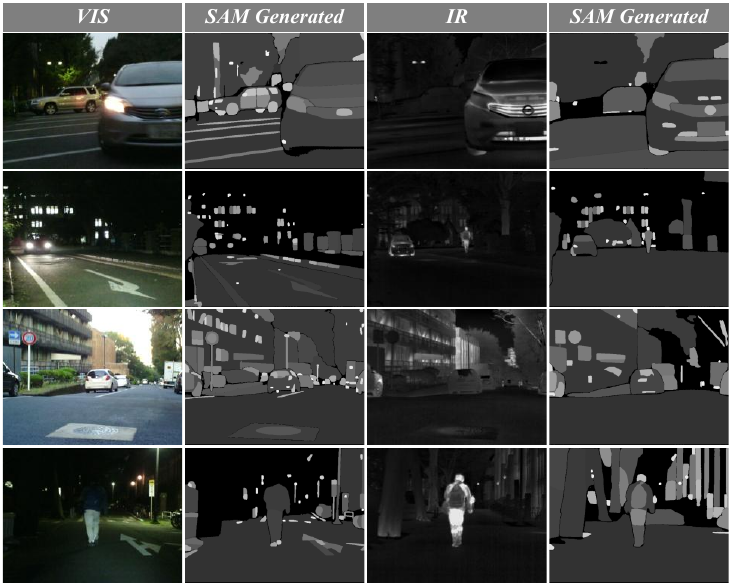}
\caption{Visible and infrared images along with their corresponding SAM semantic masks. These semantic cues are used to guide the diffusion model optimization in the second stage.}
\label{figure}
\end{figure}

Despite advancements in feature extraction and fusion strategies, inherent design limitations in existing methods lead to a core problem: a 'semantic blindness'. For instance, methods based on Convolutional Neural Networks (CNNs) are constrained by their local receptive fields, an architectural limitation that hinders their ability to comprehend global scene structures, often resulting in blurred boundaries for large-scale targets or inconsistent background luminance. On the other hand, while Generative Adversarial Networks (GANs)-based approaches can produce visually sharp images, their training relies on pixel-level fidelity constraints (e.g., intensity and gradient losses), which struggle to capture complex cross-modal semantic relationships. This not only causes training instability but also frequently leads to the erroneous suppression of salient thermal targets in infrared images. These specific design flaws point to a fundamental issue: the models are unable to semantically differentiate between foreground targets and background textures, thus failing to perform selective information enhancement. This "can't see the forest for the trees" paradigm directly causes a cascade of problems, including blurred target boundaries, loss of critical structures, and erroneous suppression of thermal features. Ultimately, this semantic deficiency is the fundamental bottleneck constraining the efficacy of fused images in downstream tasks like object detection and semantic segmentation. As illustrated in Figure 1, the failure of existing models to preserve task-critical target regions severely undermines the practical utility of fusion technology.

Recent advancements in large-scale vision models offer new promise \cite{12}. The SAM \cite{13} represents a key breakthrough owing to its outstanding semantic segmentation capabilities. The high-quality semantic masks it generates can provide multi-modal images with rich semantic priors (as shown in Figure 2). This wealth of semantic information can significantly alleviate the "semantic blindness" inherent in existing methods. It guides the model to more effectively preserve the thermal radiation features and visible details of critical targets, thereby achieving a more semantically consistent fused representation. On the other hand, image fusion, as an image generation task, is typically accompanied by processes of information compression and modal reconstruction \cite{14}. Although this process can enhance visual consistency, it often leads to a degradation in detail quality, manifesting as blurred textures, indistinct boundaries, and visual artifacts. Such degradation not only diminishes the perceptual quality of the fused image but also significantly impairs the performance of downstream tasks that rely on structural information, such as object detection and semantic segmentation. Diffusion models, by virtue of their iterative denoising generation mechanism, can stably produce high-fidelity, artifact-free images, making them exceptionally well-suited for the demands of IVIF, namely detail restoration and modal balance. More importantly, when integrated with high-quality semantic priors, diffusion models can perform semantically-aware conditional generation. This enables the fusion process to be more targeted in its representation of structures, textures, and target regions, significantly enhancing its adaptability for downstream tasks.

Building upon these insights, we argue for a conceptual shift in image fusion—moving from traditional pixel reorganization to a SGG framework. We contend that a high-fidelity fused image should be synthesized anew by a generative model, explicitly steered by high-level semantic priors. To implement this, we propose the SGDFuse Network. Recognizing the inherent task conflict between low-level feature alignment and high-level iterative generation, SGDFuse employs a two-stage decoupling design: the first stage focuses on multi-modal feature extraction to generate a robust structural prior, while the second stage leverages a conditional diffusion model for task-oriented semantic enhancement. To ensure deep integration, we introduce a holistic semantic guidance system that operates across the entire Input-Process-Output pipeline—incorporating SAM-derived masks as dense spatial constraints, modeling spatio-semantic correlations within the denoising process, and enforcing semantic consistency via a novel mask-guided loss. Experimental results (as shown in Figure 3) demonstrate that this synergy between structural understanding and detail generation significantly boosts visual quality and adaptability for downstream tasks.
\begin{figure*}[!t]
\centering
\includegraphics[scale=0.95]{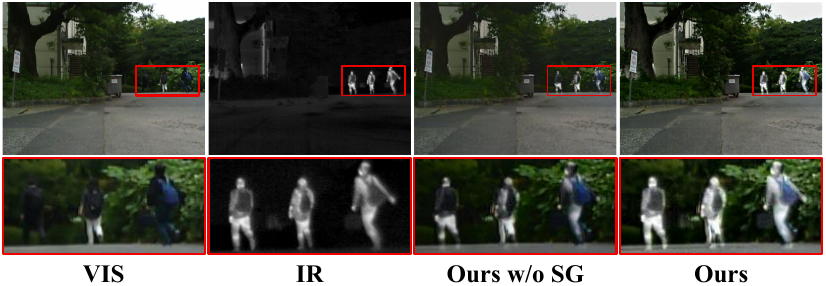}
\caption{Visual Ablation for the Necessity of the Semantic-Guided Generation Methodological Framework. The red boxes show zoomed-in detail regions.}
\label{fig2}
\end{figure*}

Our main contributions are summarized as follows:

\begin{itemize}

\item We establish a novel SGG methodological framework for image fusion. By reframing the task from traditional pixel reorganization to semantically-steered generation, this framework provides a principled and robust solution to the long-standing semantic blindness problem in the field.

\item We propose the SGDFuse, a two stage framework that serves as the first effective implementation of the SGG methodological framework. Its design strategically decouples structural understanding (Stage I) from semantic generation (Stage II), successfully resolving the inherent task conflict between cross-modal alignment and high-fidelity reconstruction.

\item We design a holistic Input-Process-Output guidance system to operationalize the SGG framework. This system reinforced by explicit SAM priors and a novel Mask-Guided Loss function ($L_{stage2}$), ensures superior preservation of key semantic information. Extensive evaluations on IVIF, medical imaging, and downstream tasks demonstrate the framework's high study significance and potential for practical implementation in complex vision systems.

\end{itemize}

\section{Related work}
\noindent 

\subsection{Deep Multimodal Image Fusion}

Recent advances in deep learning have significantly propelled multimodal image fusion, which integrates information from different sensors or viewpoints to produce clearer, higher-quality images \cite{15}. Leveraging deep neural networks’ strong feature learning and nonlinear modeling abilities, various data-driven infrared and visible image fusion methods have been developed, including those based on autoencoders (AE), CNNs, GANs, and Transformers.

AE-based methods, such as DenseFuse \cite{9}, utilize an encoder-decoder architecture for multi-scale feature extraction and fusion. While advanced variants like SEDRFuse \cite{16} introduce refinements using residual and attention mechanisms, their fundamental reliance on pixel-level loss restricts the capture of deep semantics, often leading to a loss of critical contextual information.

Methods based on CNNs enhance feature extraction. Jiang et al. \cite{17} ingeniously designed a harmonized representation space for multi-modality representation and achieve one of the best infrared and visible registration results. Li et al. \cite{18} proposed a novel multimodal medical image fusion framework that leverages semantic-level information to offer clinicians a more efficient way of observing lesions across different imaging modalities. However, their fundamental reliance on local receptive fields limits the ability to model global context and long-range dependencies, thereby compromising the semantic consistency of the fused image.

Transformer-based methods provide a new fusion paradigm, leveraging long-range dependency modeling for superior cross-domain integration. Various works have employed Transformers to enhance inter-modal dependencies, suppress redundancy, and integrate complementary features \cite{19}.  Building on this foundation, research has extended into more specialized and cutting-edge applications. Li et al. \cite{20} proposed a U-Net-based framework, pioneering the exploration of adversarial robustness in fusion tasks. To achieve precise semantic guidance, Liu et al. \cite{21} pioneered the application of prompt learning to infrared and visible image fusion, with achieving sota performance. 

GAN-based approaches introduced an unsupervised adversarial learning paradigm to image fusion. FusionGAN \cite{22} uses adversarial training between generator and discriminator. AttentionFGAN \cite{23} adds multi-scale attention to separately emphasize infrared foreground and visible background. However, despite their excellent feature preservation, these methods often suffer from training instability. Furthermore, the generator's reliance on low-level fidelity constraints (e.g., gradient, intensity) hinders the modeling of complex inter-modal distributions in the latent space, thereby compromising high-level semantic integrity.

By contrast, our method introduces semantic prior masks and a high-fidelity generation mechanism to produce fused images that are both semantically consistent and rich in detail, significantly enhancing support for downstream tasks.

\subsection{SAM and Its Applications}

The SAM \cite{13}, as a powerful pretrained model, has become a current research hotspot in the field of image processing due to its zero-shot generalization capability in image segmentation and object detection tasks \cite{24}. SAM can provide high-precision semantic masks across various vision tasks and has been widely applied in low-level vision tasks such as dehazing \cite{25}, and low-light image enhancement \cite{26}, as well as high-level tasks like object detection, demonstrating outstanding performance. By deeply understanding image content, SAM effectively captures fine details and accurately delineates target regions. Therefore, SAM is particularly suitable for infrared and visible image fusion tasks, we leverage SAM-generated segmentation masks as explicit, high-level semantic priors to guide the fusion process.

\subsection{Diffusion Models}

Diffusion models have emerged as a powerful class of deep generative models, achieving state-of-the-art results in image generation, restoration, super-resolution, and image-to-image translation tasks \cite{5}, \cite{28}. Beyond generative applications, their learned representations have also shown potential in discriminative tasks such as classification, segmentation, and object detection. A diffusion model consists of two stages: a forward process that gradually adds Gaussian noise to the input data, and a reverse process that iteratively denoises the corrupted data by predicting and removing noise, ultimately recovering high-quality samples \cite{5}.

The advantages of diffusion models lie in their ability to generate high-quality and diverse samples while effectively capturing deep underlying features of the data, which leads to outstanding performance in image generation tasks. With the continuous advancement of diffusion models across various fields, particularly in image generation, they have begun to surpass the long-standing dominance of GANs. In the context of infrared and visible image fusion, diffusion models provide powerful support for generating high-quality fused images. 

Therefore, This paper explores a diffusion model-based approach for infrared and visible image fusion, aiming to achieve superior fusion results through this generative paradigm, thereby enhancing both task performance and image quality.

\section{Methodology}
\label{section3}

\subsection{Motivation}

Existing IVIF methods face a core bottleneck in semantic understanding, as they treat fusion as a mathematical reorganization of low-level visual features (e.g., pixel intensity, gradients). This failure to semantically distinguish between critical targets and background textures acts as a fundamental obstacle, severely limiting the potential of the fused images in downstream tasks like object detection and semantic segmentation.

Our primary and core motivation is to inject explicit semantic understanding capabilities into the fusion process. We are no longer satisfied with the indiscriminate processing of low-level features by existing methods, but rather seek a novel mechanism that enables the model to truly "understand" the image and distinguish "target" from "background". For this, we introduce SAM. The high-quality semantic masks it generates can provide the fusion process with unprecedented, strong semantic priors concerning scene structure and object boundaries, which becomes our ideal tool for solving the core problem of "semantic blindness". However, high-level semantic guidance alone is insufficient. To accurately translate this precise semantic intent—for instance, to preserve the thermal features of a target while enhancing the texture of the background'—into a high-quality, pixel-level output, one must rely on an equally powerful high-fidelity generative framework. Traditional reconstruction methods are prone to generating artifacts and detail loss when handling such complex conditions, which severely undermines the efficacy of semantic guidance. Therefore, our second motivation is to employ a Diffusion Model, renowned for its superior generative capabilities, as the backbone for image reconstruction. This ensures that the high-level semantic commands can be rendered losslessly and with maximum fidelity.

The core of our Methodological Framework is the Deep Coupling of semantic understanding and image generation, rather than treating them as isolated steps. Achieving this, however, required solving two fundamental design challenges, which in turn form our core methodological innovations. First, we must resolve the inherent "task conflict" between low-level fusion and high-level generation. A naive end-to-end model struggles to simultaneously learn both. Therefore, our first core design choice is a deliberate two-stage architecture. Stage I is dedicated to generating a robust structural prior ($F_1$), which significantly simplifies the learning task for Stage II by allowing it to focus only on high-fidelity, semantic-aware refinement. Second, we must move beyond simplistic conditioning to achieve deep semantic guidance. Our second core design choice is a holistic, "Input-Process-Output" guidance system . We establish an explicit spatial prior at the Input (via concatenation), but this is merely the first step of our complete guidance loop. we use SAM masks as an explicit spatial prior at the Input, which is then deeply modeled by the denoising network (Process), and rigorously enforced at the Output by our novel Mask-Guided Loss ($L_{stage2}$). This complete system, comprising the decoupled architecture and the closed-loop guidance, is the "Deep Coupling" we seek. It endows the generation process with Precise Semantic Directivity, enabling the model to intelligently preserve thermal targets while reconstructing texture details based on semantic importance. Ultimately, SGDFuse unifies the contradictory objectives of semantic integrity and visual quality, producing fused images that are both visually compelling and highly effective for downstream tasks.

\subsection{Network Architecture}

In this section, we provide a detailed description of the proposed image fusion framework, SGDFuse. This fusion framework 
\begin{figure*}[htbp]
\centering
\includegraphics[scale=0.405]{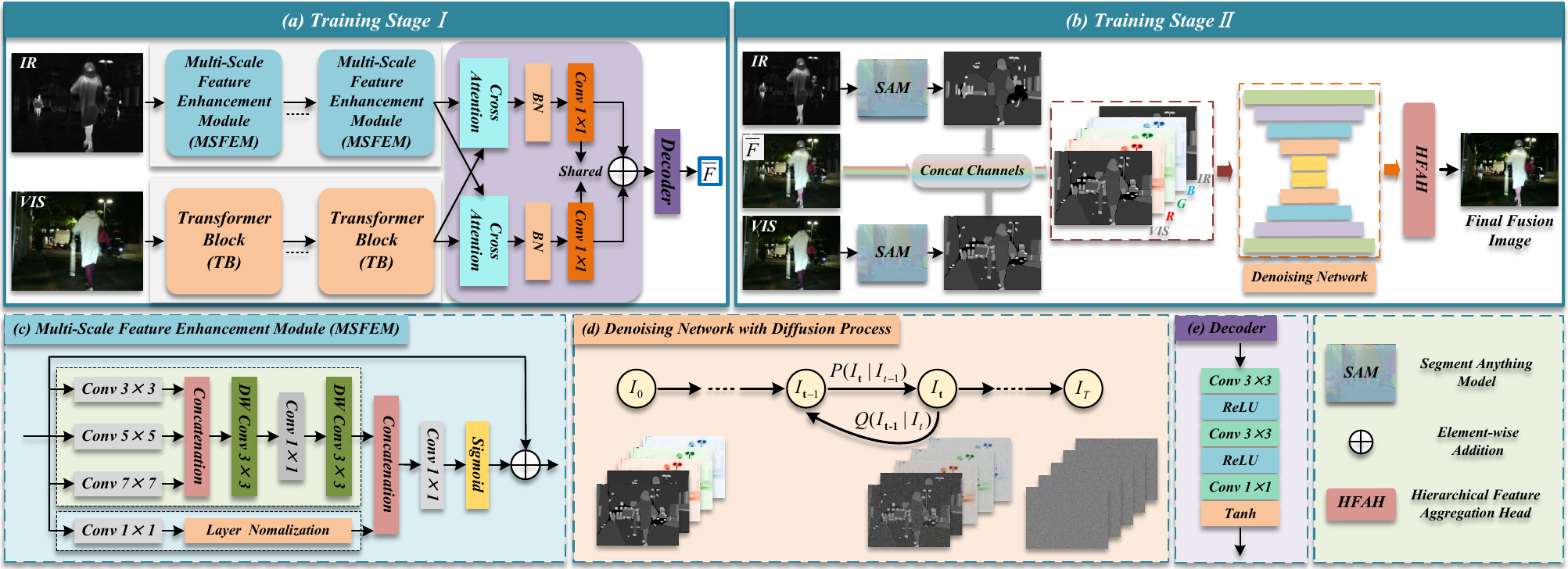}
\caption{Overall architecture of the SGDFuse framework. The framework consists of two stages: the first stage performs multi-modal feature extraction and generates an initial fused image; the second stage incorporates semantic prompts and a conditional diffusion network to refine structural details and enhance semantic consistency of the fused image.}
\label{figure}
\end{figure*}
adopts a two-stage architecture, as illustrated in Figure 4, aiming to generate high-quality fused images with structural fidelity and semantic consistency.

In Stage I, SGDFuse generates an initial fused image via a tailored multimodal feature extraction and fusion scheme. The infrared image (IR) is processed by the Multi-Scale Feature Enhancement Module (MSFEM), which uses parallel convolutional branches and channel attention to capture thermal boundaries and enhance structural cues at multiple receptive fields. Meanwhile, the visible image (VIS) is encoded by a Transformer Block (TB) with multi-head self-attention to extract global context and fine-grained texture. The extracted features are then dynamically aligned and selectively fused through a cross-attention-based interaction pathway, yielding an initial fused image that combines salient IR targets with VIS texture details.

In Stage II, SGDFuse applies a task-aware conditional diffusion mechanism to refine structure and enhance semantic consistency. Specifically, SAM generates high-quality semantic masks from the IR and VIS images, which are concatenated with the initial fused image to form a five-channel input guiding the diffusion process. A denoising decoder progressively reconstructs a high-fidelity fused image via reverse diffusion. Additionally, a Hierarchical Feature Aggregation Head (HFAH) integrates semantic features across diffusion stages, strengthening edge, structure, and region consistency, thereby improving generalization in downstream tasks.

\subsection{Multi-Scale Feature Enhancement Module (MSFEM)}

To better capture target edges and thermal regions in infrared images, SGDFuse uses a MSFEM Module in stage one to enhance structural details across scales. Since infrared images have salient targets but blurred edges and missing details, MSFEM applies multi-scale modeling with attention to strengthen key region representations.

As shown in Figure 4(c), the MSFEM employs a multi-branch parallel structure, where each branch utilizes convolutional operations with different receptive fields (e.g., 1$\times$1, 3$\times$3, 5$\times$5, 7$\times$7) to capture structural features ranging from local to medium scales. Specifically, given an input image, feature representations are first extracted separately by four sets of convolutional kernels:
\begin
{equation}
F_{i}=Conv_{j\times j}(F_{ir})(i=1,2,3,4;j=1,3,5,7).
\end
{equation}  

Then, the features from the three large-scale branches $F_{2}$, $F_{3}$ and $F_{4}$ are concatenated along the channel dimension to obtain a fused multi-scale representation:
\begin
{equation}
F_{ms}=Concat(F_{2},F_{3},F_{4}).
\end
{equation}  

To enhance feature representation, the concatenated feature $F_{ms}$ is sequentially processed through modules $DWConv_{3\times3}$, $Conv_{1\times1}$ and $DWConv_{3\times3}$ for feature enhancement and compression. This process can be expressed as follows:
\begin
{equation}
F_{enh}=DWConv_{3\times3}(Conv_{1\times1}(DWConv_{3\times3}(F_{ms}))).
\end
{equation}  

Meanwhile, the 1$\times$1 convolution branch $F_{1}$ preserves shallow-level details and linear response capabilities, providing complementary information to the main features. Finally, features $F_{ms}$ and $F_{1}$ are concatenated and fused:
\begin
{equation}
F_{cat}=Concat(F_{enh},F_{1}).
\end
{equation}  

To compress the channel dimension and unify the scale, the fused features $F_{cat}$ are passed through a $Conv_{1\times1}$ for channel reduction, followed by a Sigmoid activation function to obtain normalized response weights. Finally, MSFEM employs a residual connection by element-wise addition of the normalized weights with the original input features , producing the module’s final output. This process can be expressed as follows:
\begin
{equation}
F_{out}=F_{ir}+\sigma(Conv_{1\times1}(F_{cat})).
\end
{equation}  

The residual mechanism aids feature propagation, improves training stability, and enhances nonlinear representation. Ablation studies show that stacking MSFEM three times yields optimal fusion, confirming its effectiveness in multi-scale structural modeling and infrared feature enhancement.

\subsection{Semantic-Aware Fusion Optimization Driven by Conditional Diffusion}

To enhance structural refinement and semantic consistency of the fused image, this work introduces an optimization mechanism based on a conditional diffusion model in the second stage. Unlike existing methods that directly concatenate the original infrared and visible images, our approach concatenates the initial fused image $F_{1}\in\mathbb{R}^{\mathrm{H}\times\mathrm{W}\times3}$ generated in the first stage with two semantic masks $M_{ir}\in\mathbb{R}^{\mathrm{H}\times\mathrm{W}\times1}$ and $M_{vis}\in\mathbb{R}^{\mathrm{H}\times\mathrm{W}\times1}$ extracted from the infrared and visible images using the SAM, forming a five-channel task-aware input. We employ the Denoising Diffusion Probabilistic Model (DDPM) \cite{30} to model the diffusion generation process of this five-channel input in the latent space. The forward process gradually adds Gaussian noise over T time steps to produce a perturbed data distribution, while the reverse process progressively denoises the input under the guidance of semantic masks to reconstruct a fused image with clear structure and semantic consistency. By modeling the joint distribution of the fused image and semantic priors, the diffusion model learns a semantic-aware generative pathway, thereby further improving fusion quality and adaptability to downstream tasks.

\begin{figure}[htbp]
\centering
\includegraphics[scale=0.75]{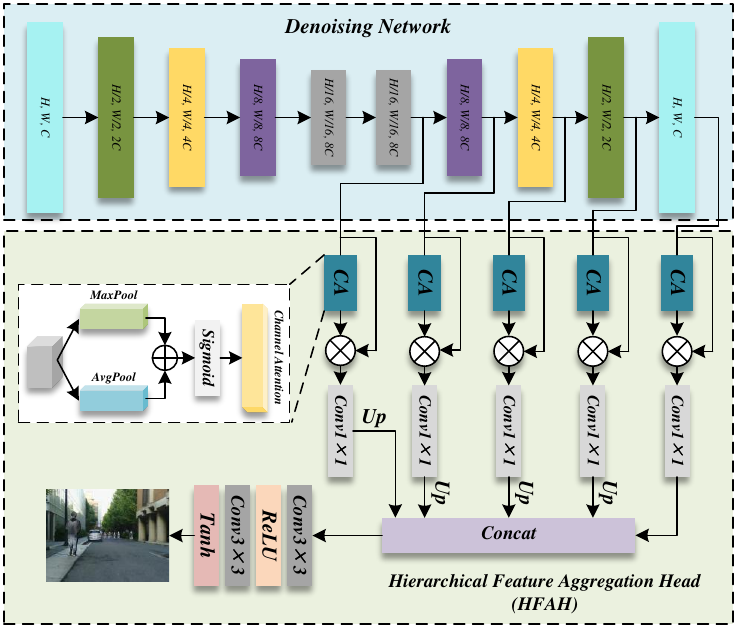}
\caption{Architecture of the denoising network and the Hierarchical Feature Aggregation Head (HFAH).}
\label{figure}
\end{figure}
\subsubsection{Forward Diffusion Process} The diffusion process consists of a forward perturbation process (Forward Process) and a reverse generation process (Reverse Process), whose core idea is derived from non-equilibrium thermodynamics theory \cite{31} and is systematically modeled in DDPM \cite{30}. In the forward diffusion process, the five-channel input $\mathrm{I}_{0}\in\mathbb{R}^{\mathrm{H}\times\mathrm{W}\times5}$, formed by concatenating the initial fused image and semantic masks, is gradually perturbed by Gaussian noise through a Markov chain of T time steps, causing its distribution to progressively approach a standard Gaussian distribution. Specifically, the noisy image $I_{t}$ at the t-th time step can be represented by the following conditional distribution:
\begin
{equation}
P(I_t|I_{t-1})=\mathcal{N}(I_t;\sqrt{\alpha_t}I_{t-1},(1-\alpha_t)\mathbf{I}),
\end
{equation}  
where $\alpha_{\mathrm{t}}\in(0,1)$ denotes the variance scheduling factor at the t-th time step, $\mathbf{I}$ represents the identity covariance matrix, and the Gaussian noise follows a standard normal distribution.

Furthermore, to improve sampling efficiency, the model supports directly generating the corresponding noisy image $I_{t}$ at any arbitrary time step t based on the original input $I_{0}$. The calculation is formulated as follows:
\begin
{equation}
I_t=\sqrt{\bar{\alpha}_t}I_0+\sqrt{1-\bar{\alpha}_t}\varepsilon,\quad\varepsilon\sim\mathcal{N}(0,\mathbf{I}),
\end
{equation}  
where $\overline{\alpha}_{t}=\prod_{i=1}^{t}\alpha_{i}$ denotes the cumulative noise retention factor up to time step t, and $\varepsilon$ is a noise variable sampled from the standard Gaussian distribution.

In this study, the five-channel input is constructed by concatenating the three-channel preliminary fused image $F_{1}\in\mathbb{R}^{\mathrm{H}\times\mathrm{W}\times3}$ with two semantic masks $M_{ir}\in\mathbb{R}^{\mathrm{H}\times\mathrm{W}\times1}$ and $M_{vis}\in\mathbb{R}^{\mathrm{H}\times\mathrm{W}\times1}$ generated by the SAM:
\begin
{equation}
I_0=\mathrm{Concat}(F_1,M_{ir},M_{vis})\in\mathbb{R}^{H\times W\times5}.
\end
{equation}  

This design enables the diffusion process to model not only the visual appearance of the image but also incorporate task-relevant semantic guidance, thereby effectively enhancing structural preservation and semantic consistency during the subsequent reverse generation process.

\subsubsection{Reverse Diffusion Process} During the reverse diffusion process, the diffusion model employs a parameterized neural network to progressively denoise the noisy image at each time step, ultimately restoring a high-quality fused image relevant to the task. Specifically, at each time step $t$, the network estimates the current noisy image $I_{t}\in\mathbb{R}^{\mathrm{H}\times\mathrm{W}\times5}$ to obtain the image at the previous time step $I_{t-1}$. This process can be modeled as a conditional Gaussian distribution:
\begin
{equation}
Q(I_{t-1}|I_t)=\mathcal{N}(I_{t-1};\mu_\theta(I_t,t),\sigma_t^2\mathbf{I}),
\end
{equation} 
where $\mu_\theta(I_t,t)$ denotes the mean predicted by the neural network at the current time step $t$, and $\sigma_t^2$ represents the noise variance at this time step, which is expressed as follows:
\begin
{equation}
\sigma_t^2=\frac{1-\bar{\alpha}_{t-1}}{1-\bar{\alpha}_t}\cdot\beta_t,
\end
{equation} 
where $\beta_{t}=1-\alpha_{t}$ denotes the noise level added at each step, and $\bar{\alpha}_t$ is the cumulative noise retention factor up to step t. The mean term $\mu_\theta(I_t,t)$ of the conditional distribution can be expressed as follows:
\begin
{equation}
\mu_\theta(I_t,t)=\frac{1}{\sqrt{\alpha_t}}\left(I_t-\frac{\beta_t}{\sqrt{1-\bar{\alpha}_t}}\cdot\varepsilon_\theta(I_t,t)\right),
\end
{equation} 
where $\varepsilon_\theta(I_t,t)$ represents the noise predicted by the denoising network (U-Net) for the current image $I_{t}$ at time step t.

In the proposed method, the input image $I_{t}\in\mathbb{R}^{\mathrm{H}\times\mathrm{W}\times5}$ consists of the three-channel initial fused image concatenated with two semantic masks. Therefore, the reverse diffusion process not only restores detailed structures but also enhances semantic consistency under the guidance of semantic prompts. By iteratively performing this process, the diffusion model progressively reconstructs a high-fidelity and semantically coherent final fused image from standard Gaussian noise.

\subsubsection{Loss Function of Diffusion Process} In this work, we adopt the simplified training objective consistent with DDPM \cite{30}, which aims to minimize the mean squared error between the noise predicted by the model and the actual added Gaussian noise. The loss function is expressed as follows:
\begin
{equation}
\mathcal{L}_{\mathrm{diff}}=\left\|\gamma-\varepsilon_\theta\left(\sqrt{\bar{\alpha}_t}I_0+\sqrt{1-\bar{\alpha}_t}\gamma,t\right)\right\|_2^2,
\end
{equation} 
where $I_{t}\in\mathbb{R}^{\mathrm{H}\times\mathrm{W}\times5}$ denotes the original five-channel input image formed by concatenating the preliminary fused image and semantic masks; $\gamma\in\mathrm{N}(0,\mathrm{I})$ represents the noise sampled from a standard normal distribution; ${\bar{\alpha}_t}$ is the cumulative retention coefficient up to time step $t$ and $\varepsilon_{\theta}$ is the parameterized denoising network. By minimizing this loss, the model learns to effectively recover the clean image from noisy inputs at any time step, enabling a semantic-aware denoising generation process.

\subsubsection{Structure of the Denoising Network} In the second stage of our framework, to accurately estimate the Gaussian noise added during the forward diffusion process, we design a denoising network $\varepsilon_{\theta}(\cdot,\cdot)$ based on a U-Net architecture. Inspired by the design principles of SR3 \cite{31}, the proposed network exhibits strong image restoration capability and multi-scale representation learning, enabling effective modeling of the latent structure within the semantically guided five-channel input. Specifically, the denoising network comprises three main components: a contracting path (consisting of five convolutional layers) that progressively downsamples the input to extract deep semantic and structural features; an expanding path (with five corresponding convolutional layers) that incrementally restores spatial resolution; and a diffusion head that outputs noise predictions matching the input dimensions. Leveraging the five-channel input formed by concatenating the preliminary fused image and semantic masks, the network is capable of capturing both low-level texture and edge structures as well as high-level semantic cues, thereby facilitating semantic-aware high-fidelity image reconstruction.

\subsection{Hierarchical Feature Aggregation Head (HFAH)}

To further enhance feature representation in the decoding stage, we introduce a HFAH into the decoder path of the denoising network, as shown in Figure 5. HFAH integrates multi-level decoded features and incorporates a spatial attention mechanism to jointly optimize structural detail and semantic consistency. Specifically, all weighted hierarchical features are concatenated and passed through a fusion head to generate the final fused image $I_{f}\in\mathbb{R}^{\mathrm{H}\times\mathrm{W}\times3}$, which exhibits clearer structural boundaries and improved detail preservation. The fusion head consists of multiple 3$\times$3 convolutional layers designed to map the high-dimensional, multi-scale diffusion features into a final three-channel fused image. A Tanh activation function is applied at the output to enhance texture continuity and fine detail expression. Benefiting from HFAH's precise modeling of structural edges and semantic regions, this module significantly improves the boundary fidelity and fine-grained reconstruction quality of the final fused image.

\subsection{Loss Function of Training Process}

To achieve semantically consistent and structurally clear multimodal image fusion, this paper designs task-specific loss functions under the two-stage training framework of SGDFuse. Specifically, dedicated loss objectives are formulated for the structural representation of the preliminary fused image in the first stage, and for the high-quality generation of the final fused image in the second stage.

\subsubsection{Stage I: Supervision of Preliminary Fusion} The first stage aims to achieve modal alignment and information fusion between infrared and visible images, producing a preliminary fused image $F_{1}\in\mathbb{R}^{\mathrm{H}\times\mathrm{W}\times3}$. Gradient Loss and Intensity Loss are employed in this stage to ensure structural consistency with the visible image and preserve thermal information from the infrared image, defined as:
\begin
{equation}
L_{grad}^{1}=\frac{1}{HW}\parallel\nabla F_{1}-\nabla I_{vis}\parallel,
\end
{equation} 
\begin
{equation}
L_{int}^{1}=\frac{1}{HW}\parallel F_{1}-I_{ir}\parallel,
\end
{equation} 
where $\nabla$ denotes the image gradient operator used to extract edge structures, H and W are the height and width of the image.

The overall loss for the first stage is defined as:
\begin
{equation}
L_{stage1}=L_{int}^{1}+L_{grad}^{1},
\end
{equation} 

\subsubsection{Stage II: Diffusion Optimization Training} In the second stage, the preliminary fused image is concatenated with the infrared and visible semantic masks obtained from SAM to form a five-channel input, which guides the conditional diffusion model to generate the final fused image $I_{f}\in\mathbb{R}^{\mathrm{H}\times\mathrm{W}\times3}$. To further enhance structural fidelity and semantic consistency, the fusion process is supervised using the following loss functions.

Mask-guided Intensity Loss: To enhance the luminance consistency and thermal response representation of the fused image in semantically salient regions, a region-weighted intensity loss is introduced. Let the infrared and visible semantic masks generated by SAM be denoted as $M_{ir}\in\mathbb{R}^{\mathrm{H}\times\mathrm{W}\times1}$ and $M_{vis}\in\mathbb{R}^{\mathrm{H}\times\mathrm{W}\times1}$, respectively. A joint region mask $M\in\mathbb{R}^{\mathrm{H}\times\mathrm{W}\times1}$ is constructed as follows:
\begin
{equation}
\mathrm{M=max(M_{ir},M_{vis})},
\end
{equation} 
\begin
{equation}
L_{int}^{mask}=\frac{1}{HW}\parallel M\cdot(I_{f}-max(I_{ir},I_{vis}))\parallel_{1},
\end
{equation} 
where $I_{f}\in\mathbb{R}^{\mathrm{H}\times\mathrm{W}\times3}$ denotes the final fused image.

Mask-guided Gradient Loss: To further enhance edge clarity within salient regions, a region-weighted gradient loss is introduced. This loss guides the fused image to align with the clearer edges from either the infrared or visible image within the salient areas. It is defined as follows:
\begin
{equation}
L_{grad}^{mask}=\frac{1}{HW}\parallel M\cdot(\nabla I_{f}-max(\nabla I_{ir},\nabla I_{vis}))\parallel_{1},
\end
{equation} 
where $\nabla$ denotes the image gradient operator used to extract edge structures; the loss is applied within the masked regions to guide the network to focus on optimizing the structural edges of the target areas.

Finally, the total loss for the second stage is defined as:
\begin
{equation}
L_{stage2}=\lambda_{1}\cdot L_{int}^{mask}+\lambda_{2}\cdot L_{grad}^{mask},
\end
{equation} 
where $\lambda_{1}$ and $\lambda_{2}$ are weighting coefficients that control the relative importance of regional intensity alignment and edge alignment. Based on experimental validation, the optimal values are determined as $\lambda_{1}$=1.5 and $\lambda_{2}$=1.

\section{EXPERIMENTS}

\subsection{Experimental Settings}

\subsubsection{Datasets} To comprehensively evaluate the effectiveness and applicability of the proposed model, we selected four datasets for experimentation: MSRS \cite{10}, LLVIP \cite{32}, M$^3$FD \cite{33}, and RoadScene \cite{1}. The MSRS dataset provides 361 pairs of test images with a resolution of 640$\times$480; the M$^3$FD dataset contains 4,164 image pairs with a resolution of 1024$\times$768; the LLVIP dataset includes 16,836 image pairs with a resolution of 1280$\times$1024. The RoadScene dataset consists of 221 registered pairs of infrared and visible images.

\subsubsection{Evaluation Metrics} We employ seven commonly used metrics for quantitative evaluation. Specifically, these include Entropy (EN), Standard Deviation (SD), Spatial Frequency (SF), Mutual Information (MI), Spectral Color Distortion (SCD), and Visual Information Fidelity (VIF). Higher values of these metrics indicate better image quality.

\subsubsection{Training Details} The proposed model is trained on the MSRS dataset, comprising 1,083 visible-infrared image pairs for training and 361 pairs for testing. Following \cite{5}, we crop 160$\times$160 image patches during training to enhance data diversity. Inspired by \cite{34}, multi-channel diffusion features are extracted across multiple time steps to enrich representations. The fusion network is trained for 200 epochs using the Adam optimizer \cite{35} with a learning rate of 1e-4 and batch size of 24. During testing, the model processes input images at their original resolution, outputting a three-channel color fused image suitable for subsequent qualitative and quantitative evaluations. The entire model is implemented using the PyTorch framework, and all experiments are conducted on a workstation equipped with an NVIDIA RTX 4090 GPU.

\subsection{Generalization Comparison Experiments}

To evaluate the generalization capability of our network, we compare it against the following representative image fusion methods: U2Fusion \cite{1}, TarDAL \cite{33}, PIAFusion \cite{10}, IGNet \cite{36}, CDDFuse \cite{37}, Dif-fusion \cite{5}, PFCFuse \cite{38}, SHIP \cite{24}, EMMA \cite{28}, Text-DiFuse \cite{11}, DAFusion \cite{15}, MaeFuse \cite{6}, and SAGE \cite{12}. These SOTA baselines cover a wide range of architectures, including recent diffusion-based models (e.g., Dif-fusion \cite{5}, Text-DiFuse \cite{11}) and state-of-the-art methods based on Transformer architectures (e.g., MaeFuse \cite{6} and PFCFuse \cite{38}).
\begin{table*}[t]
\centering
\caption{Quantitative comparison on MSRS and M$^3$FD datasets, with red representing the best results, blue representing the second best results.}
\resizebox{1\textwidth}{!}{
\begin{tabular}{l|ccccccc|ccccccc}
\toprule
\multirow{2}{*}{\textbf{Method}} & \multicolumn{7}{c|}{\textbf{MSRS}} & \multicolumn{7}{c}{\textbf{M$^3$FD}} \\
\cmidrule(lr){2-8} \cmidrule(lr){9-15}
& \textbf{EN} & \textbf{SD} & \textbf{SF} & \textbf{MI} & \textbf{SCD} & \textbf{VIF} & \textbf{$Q^{abf}$} & \textbf{EN} & \textbf{SD} & \textbf{SF} & \textbf{MI} & \textbf{SCD} & \textbf{VIF} & \textbf{$Q^{abf}$}\\
\midrule
\textbf{U2Fusion}\cite{1} & 5.37 & 25.52 & 9.07 & 1.4 & 1.24 & 0.51 & 0.42 & 6.28 & 26.26 & 8.36 & 1.79 & 1.22 & 0.68 & 0.48\\
\textbf{TarDAL}\cite{33} & 5.28 & 25.22 & 5.98 & 1.49 & 0.71 & 0.44 & 0.18 & 6.81 & 38.57 & 6.78 & 2.27 & 1.21 & 0.60 & 0.31\\
\textbf{PIAFusion} \cite{10} & 6.64 & {\color[HTML]{FF0000} 45.34} & 12.12 & 2.75 & 1.7 & 1.02 &0.67 & 6.87 & 34.01 & 12.49 & 2.61 & 1.43 & 0.94 & 0.60\\
\textbf{IGNet}\cite{36} & 6.12 & 33.91 & 9.66 & 1.29 & 1.56 & 0.69 &0.47 & 6.94 & {\color[HTML]{FF0000} 39.93} & 12.17 & 1.45 & 1.54 & 0.74 & 0.45 \\
\textbf{CDDFuse} \cite{37} & 6.70 & 43.38 & 11.55 & {\color[HTML]{FF0000} 3.43} & 1.62 & 1.05 & 0.69 & 6.85 & 37.47 & 12.87 & 2.27 & 1.60 & 0.90 & 0.61\\
\textbf{Dif-fusion}\cite{5} & 6.47 & 37.69 & 10.86 & 1.78 & 1.47 & 0.82 &0.53 & 6.59 & 32.71 & 11.41 & 2.01 & 1.19 & 0.48 &0.42\\
\textbf{PFCFuse} \cite{38} & 6.69 & 43.12 & 11.60 & 2.65 & 1.63 & 0.92 &0.61 & 6.85 & 37.62 & {\color[HTML]{FF0000} 13.29} & 2.15 & 1.55 & 0.69 &0.51\\
\textbf{SHIP} \cite{24} & 6.43 & 41.13 & 11.81 & 2.86 & 1.51 & 1.04 &0.66 & 6.63 & 32.46 & 13.45 & 2.91 & 1.11 & 0.86 & 0.62\\
\textbf{EMMA} \cite{28} & 6.71 & 44.13 & 11.56 & 2.94 & 1.63 & 0.97 &0.64 & 6.93 & {\color[HTML]{0000FF}38.31} & 12.84 & 2.79 & 1.49 & 0.66 & 0.58 \\
\textbf{Text-Difuse}\cite{11} & 6.65 & 44.96 & 12.97 & 2.89 & 1.45 & 0.84 &0.69 & 6.91 & 37.78 & 13.08 & 2.91 & 1.63 & 0.95 & 0.61\\
\textbf{DAFusion} \cite{15} & 6.69 & 45.02 & 12.84 & 2.56 & 1.62 & 0.91 & 0.68 & 6.94 & 37.78 & 13.08 & 2.91 & 1.63 & 0.95 &0.63\\
\textbf{MaeFuse} \cite{6} & {\color[HTML]{0000FF} 6.74} & 45.19 & 12.96 & 2.97 & 1.69 & {\color[HTML]{0000FF} 1.06} &{\color[HTML]{0000FF} 0.72}& {\color[HTML]{0000FF}6.96} & 37.98 & 13.24 & {\color[HTML]{0000FF}2.97} & {\color[HTML]{0000FF}1.68} & {\color[HTML]{0000FF}1.00} &{\color[HTML]{0000FF}0.66}\\
\textbf{SAGE }\cite{12} & 6.71 & 45.04 & {\color[HTML]{0000FF} 13.17} & 2.91 & {\color[HTML]{0000FF} 1.71} & 1.03 & 0.70 & 6.87 & 37.86 & 13.15 & 2.94 & 1.65 & 0.97 & 0.64 \\
\midrule
\textbf{SGDFuse(Ours)} & {\color[HTML]{FF0000} 6.81} & {\color[HTML]{0000FF} 45.28} & {\color[HTML]{FF0000} 13.27} & {\color[HTML]{0000FF} 2.99} & {\color[HTML]{FF0000} 1.73} & {\color[HTML]{FF0000} 1.08} &{\color[HTML]{FF0000} 0.74}& {\color[HTML]{FF0000}6.99} & 38.02 & {\color[HTML]{0000FF}13.27} & {\color[HTML]{FF0000}2.99} & {\color[HTML]{FF0000}1.69} & {\color[HTML]{FF0000}1.01} & {\color[HTML]{FF0000}0.69} \\
\bottomrule
\end{tabular}}
\end{table*}

\begin{figure*}[t]
\centering
\includegraphics[scale=0.505]{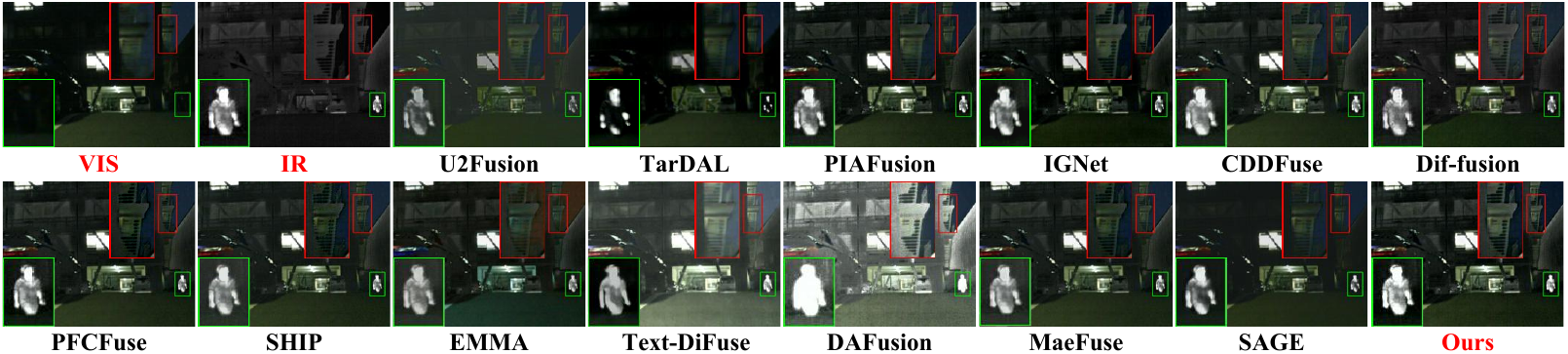}
\caption{Qualitative comparisons between our method and thirteen state-of-the-art fusion approaches on the MSRS dataset.}
\label{figure}
\end{figure*}
\begin{figure*}[t]
\centering
\includegraphics[scale=0.505]{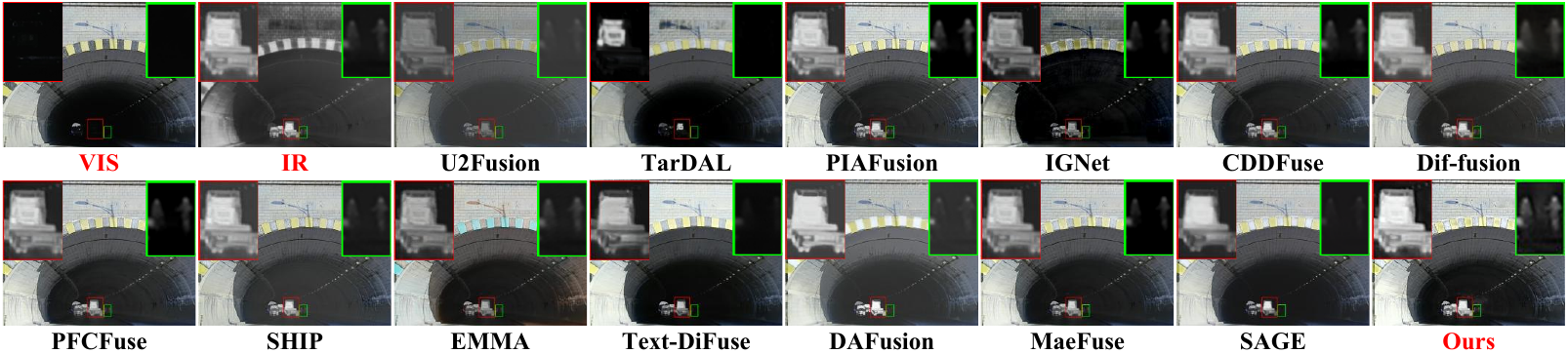}
\caption{Qualitative comparisons between our method and thirteen state-of-the-art fusion approaches on the M$^3$FD dataset.}
\label{figure}
\end{figure*}

\subsubsection{Results and Analysis on the MSRS Dataset} As shown in Table I, our method achieves the best performance across five key metrics: EN, SD, SF, VIF, and $Q^{abf}$. The higher EN and MI values indicate better information preservation from both modalities, while superior SD and SF reflect enhanced contrast and edge sharpness. Notably, the highest $Q^{abf}$ further confirms the effectiveness of our method in maintaining structural consistency and perceptual quality.

Figure 6 shows qualitative comparisons under diverse conditions, including varying illumination, multiple targets, and day/night scenes. While U2Fusion and PLAFusion exhibit blurred details and brightness inconsistency, IGNet, CDDFuse, and MaeFuse partially enhance thermal targets but often lack texture fidelity or introduce artifacts. In contrast, our method effectively preserves thermal saliency, enhances structural clarity, and maintains global luminance balance, achieving superior perceptual quality and semantic consistency.

\subsubsection{Results and Analysis on the M$^3$FD Dataset} As shown in Table I, our method achieves the best performance across five key metrics: EN, MI, SCD, VIF, and $Q^{abf}$. The high EN (6.99) and MI scores indicate richer information content and stronger source correlation. The superior SCD reflects better structural preservation, while the leading VIF and $Q^{abf}$ scores demonstrate enhanced visual quality and edge consistency.

Figure 7 shows qualitative comparisons on challenging scenes from the M$^3$FD dataset, including nighttime roads, uneven lighting, low-contrast IR targets, and complex textures. Methods like U2Fusion and TarDAL suffer from blurred structures and missing textures, while IGNet, PLAFusion, and CDDFuse offer partial improvements but still lack edge continuity and luminance balance. MaeFuse and SAGE improve global coherence but exhibit limitations in detail preservation. In contrast, our method consistently restores thermal contours and visible textures with natural brightness and clear contrast. Especially in scenes with weak IR signals and cluttered backgrounds, SGDFuse achieves accurate foreground–background separation, highlighting its superiority in structural refinement and semantic consistency.
\begin{table*}[htbp]
\centering
\caption{Quantitative comparison on LLVIP and RoadScene datasets, with red representing the best results, blue representing the second best results.}
\resizebox{1\textwidth}{!}{
\begin{tabular}{l|ccccccc|ccccccc}
\toprule
\multirow{2}{*}{\textbf{Method}} & \multicolumn{7}{c|}{\textbf{LLVIP}} & \multicolumn{7}{c}{\textbf{RoadScene}} \\
\cmidrule(lr){2-8} \cmidrule(lr){9-15}
& \textbf{EN} & \textbf{SD} & \textbf{SF} & \textbf{MI} & \textbf{SCD} & \textbf{VIF} & \textbf{$Q^{abf}$}& \textbf{EN} & \textbf{SD} & \textbf{SF} & \textbf{MI} & \textbf{SCD} & \textbf{VIF} & \textbf{$Q^{abf}$}\\
\midrule
\textbf{U2Fusion}\cite{1} & 5.33 & 18.05 & 6.73 & 1.52 & 0.61 & 0.45 & 0.23 & 6.59 & 38.11 & 10.73 & 1.75 & 1.68 & 0.53 & 0.46 \\
\textbf{TarDAL}\cite{33} & 6.01 & 27.44 & 7.02 & 1.63 & 0.83 & 0.55 & 0.22 & 7.15 & 47.68 & 10.10 & 2.18 & 1.56 & 0.60 & 0.40 \\
\textbf{PIAFusion} \cite{10} & 7.03 & 42.88 & 16.10 & 2.26 & 1.54 & 0.91 & 0.65 & 6.99 & 44.23 & 11.51 & 2.47 & 1.58 & 0.69 & 0.46 \\
\textbf{IGNet}\cite{36} & 6.88 & 42.95 & 13.95 & 1.44 & 1.53 & 0.78 & 0.50 & 7.23 & 42.47 & 9.72 & 1.38 & 1.50 & 0.65 & 0.47 \\
\textbf{CDDFuse} \cite{37} & 7.01 & 42.47 & 14.52 & {\color[HTML]{FF0000}2.89} & 1.56 & 0.93 & 0.63 & {\color[HTML]{FF0000}7.44} & {\color[HTML]{FF0000}54.58} & {\color[HTML]{FF0000}16.29} & 2.30 & {\color[HTML]{0000FF}1.80} & 0.69 & 0.52 \\\
\textbf{Dif-fusion}\cite{5} & 7.06 & 43.30 & 14.48 & 2.02 & 1.47 & 0.72 & 0.52 & 7.10 & 44.90 & 10.11 & 2.19 & 1.52 & 0.52 & 0.41\\
\textbf{PFCFuse} \cite{38} & 6.87 & 40.62 & 15.26 & 2.62 & 1.52 & 0.90 & 0.63 & {\color[HTML]{0000FF}7.41} & 50.37 & 13.81 & 2.37 & 1.78 & 0.74 & 0.55\\
\textbf{SHIP} \cite{24} & 6.89 & 41.71 & 15.84 & 2.63 & 1.39 & 0.92 & 0.66 & 7.15 & 44.67 & 14.59 & {\color[HTML]{0000FF}2.83} & 1.41 & 0.71 & 0.58 \\
\textbf{EMMA} \cite{28} & 6.74 & 42.68 & 14.75 & 2.41 & 1.37 & 0.87 & 0.59 & 7.14 & 48.73 & 10.78 & 2.76 & 1.49 & 0.68 & 0.54\\
\textbf{Text-Difuse}\cite{11} & 6.99 & 42.94 & 15.73 & 2.58 & 1.54 & 0.90 & 0.65 & 7.21 & 50.29 & 9.49 & 1.74 & 1.24 & 0.77 & 0.37 \\
\textbf{DAFusion} \cite{15} & 7.04 & 43.13 & 16.01 & 2.64 & 1.57 & 0.89 & 0.66 & 7.19 & 50.37 & 11.59 & 1.75 & 1.69 & 0.78 & 0.57 \\
\textbf{MaeFuse} \cite{6} & 7.08 & {\color[HTML]{0000FF}43.24} & {\color[HTML]{0000FF}16.13} & 2.72 & {\color[HTML]{0000FF}1.62} & {\color[HTML]{0000FF}0.96} & {\color[HTML]{0000FF}0.69} & 7.26 & 51.64 & 14.96 & 2.72 & 1.72 & {\color[HTML]{0000FF}0.79} & {\color[HTML]{0000FF}0.60}\\
\textbf{SAGE }\cite{12} & {\color[HTML]{0000FF}7.09} & 43.19 & 16.09 & 2.67 & 1.59 & 0.93 & 0.67 & 7.21 & 51.24 & 14.73 & 2.59 & 1.79 & 0.75 & 0.59\\
\midrule
\textbf{SGDFuse(Ours)} & {\color[HTML]{FF0000} 7.12} & {\color[HTML]{FF0000} 43.35} & {\color[HTML]{FF0000} 16.15} & {\color[HTML]{0000FF} 2.79} & {\color[HTML]{FF0000} 1.65} & {\color[HTML]{FF0000} 0.99} &{\color[HTML]{FF0000} 0.71} & 7.29 & {\color[HTML]{0000FF}51.79} & {\color[HTML]{0000FF}15.68} & {\color[HTML]{FF0000}2.89} & {\color[HTML]{FF0000}1.84 }& {\color[HTML]{FF0000}0.82} & {\color[HTML]{FF0000}0.62} \\
\bottomrule
\end{tabular}}
\end{table*}

\subsubsection{Results and Analysis on the LLVIP Dataset} As shown in Table II, our proposed method attains the highest scores across all evaluation metrics except MI, where it ranks second, demonstrating strong competitiveness. These results highlight SGDFuse’s clear superiority in preserving information content, texture details, edge sharpness, and perceptual quality. Notably, leading performance in EN, SD, and SF indicates richer information, enhanced contrast, and improved clarity in the fused images. Furthermore, the superior $Q^{abf}$ and VIF scores confirm that SGDFuse achieves better perceptual consistency and edge alignment, closely aligning with human visual preferences.

The LLVIP dataset mainly contains nighttime scenes with challenges such as strong occlusion, weak textures, and extremely low illumination, which place higher demands on fusion robustness. Figure 8 presents visual comparisons under these conditions. Methods like U2Fusion and TarDAL fail to enhance thermal saliency and suffer from blurred details and suppressed backgrounds. SHIP and SAGE improve thermal contrast but struggle with texture reconstruction and luminance balance. In contrast, our method preserves IR thermal responses while effectively integrating VIS textures and structures, producing results with sharper edges, consistent brightness, and stronger global contrast. Even under severe low-light conditions, SGDFuse demonstrates high stability and perceptual quality, confirming its robustness and practical applicability.
\begin{figure*}[t]
\centering
\includegraphics[scale=0.505]{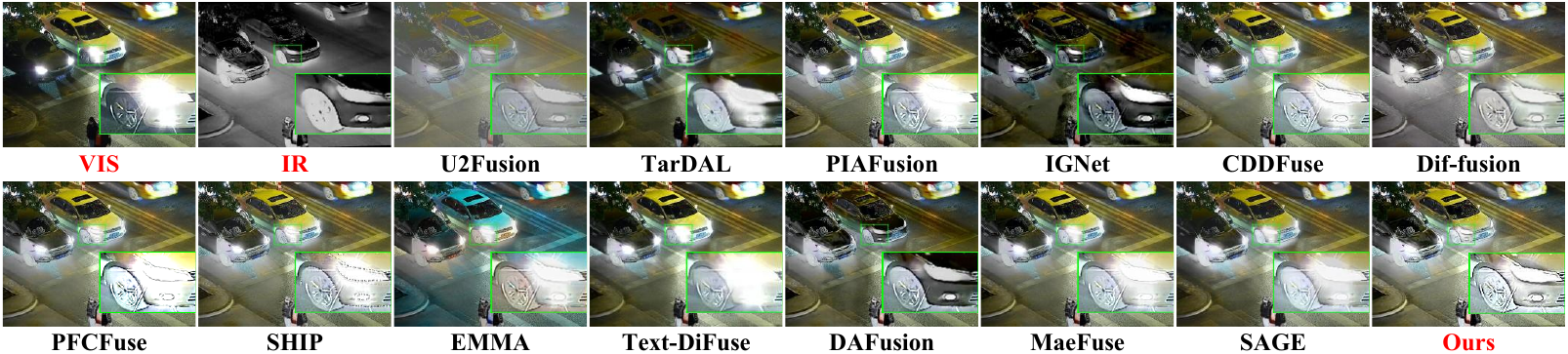}
\caption{Qualitative comparisons between our method and thirteen state-of-the-art fusion approaches on the LLVIP dataset.}
\label{figure}
\end{figure*}
\begin{figure*}[htbp]
\centering
\includegraphics[scale=0.505]{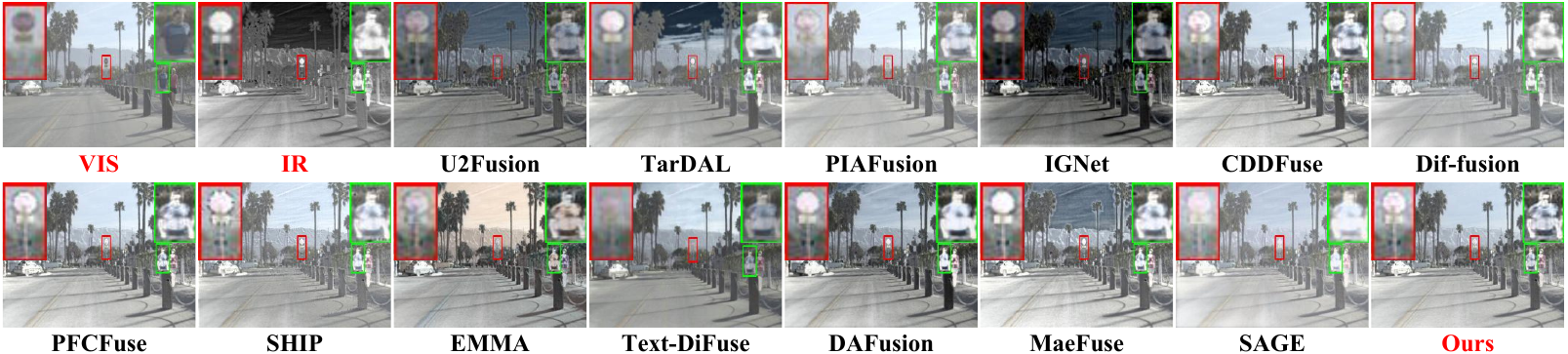}
\caption{Qualitative comparisons between our method and thirteen state-of-the-art fusion approaches on the RoadScene dataset.}
\label{figure}
\end{figure*}

\subsubsection{Results and Analysis on the RoadScene Dataset}  

As shown in Table II, our method attains the best performance on four key metrics: MI, SCD, VIF, and $Q^{abf}$. The leading MI score evidences effective preservation of complementary information from infrared and visible inputs. Superior SCD highlights strong structural and contour fidelity. Additionally, top VIF and $Q^{abf}$ scores confirm enhanced perceptual quality and edge alignment, demonstrating the method’s robust capability to produce semantically consistent and visually faithful fusion results.

The RoadScene dataset includes complex traffic scenarios with varying illumination and diverse target distributions, imposing high demands on fusion for target detection and scene understanding. Figure 9 shows visual comparisons. Methods such as U2Fusion and IGNet fail to retain both thermal cues and visual details, leading to blurred textures or weakened targets. Text-DiFuse and SAGE show limited target enhancement and detail clarity. MaeFuse preserves some textures but often produces dark results, reducing visual quality. In contrast, our method achieves superior brightness balance, structural clarity, and background suppression, demonstrating strong robustness and practical value under complex conditions.
\begin{table}[htbp]
\centering
\caption{Comparison of model parameters, latency, and FLOPs with other methods. Our latency is reported at our optimal T=60 steps. Latency for non-iterative methods is for a single forward pass.}
\label{tab:model_comparison_transposed}
\renewcommand{\arraystretch}{0.95} % 调整行高
\footnotesize % <--- 在这里添加 \footnotesize
\begin{tabular}{lccc} 
\toprule
\textbf{Method} & \textbf{Parameters(M)} & \textbf{FLOPs(G)} & \textbf{Latency(ms)} \\
\midrule
U2Fusion & 0.66 & 517.34 & 104 \\
TarDAL & 0.30 & 221.33 & 41 \\
PIAFusion & 1.50 & 529.73 & 85 \\
IGNet & 7.87 & 51.49 & 98 \\
CDDFuse & 2.00 & 36.35 & 120 \\
Diffusion & 45.37 & 376.68 & 465 \\
PFCFuse & 0.64 & 302.75 & 35 \\
SHIP & 1.56 & 340.14 & 54 \\
EMMA & 1.52 & 106.41 & 64 \\
Text-Difuse & 121.51 & 243.68 & 350 \\
DAFusion & 0.28 & 8.24 & 132 \\
MaeFuse & 35.44 & 68.27 & 78 \\
SAGE & 0.14 & 52.06 & 25 \\
\midrule[0.5pt] 
\textbf{Ours} & 47.16 & 220.74 & 59 \\
\bottomrule
\end{tabular}
\end{table}

\subsection{Practical Efficiency Analysis}

\subsubsection{Computational Complexity and Inference Efficiency}  

To comprehensively evaluate the practical inference efficiency of the proposed framework, we conducted a rigorous comparison against other SOTA methods across three dimensions: parameter count (Parameters), computational cost (FLOPs), and average single-image inference latency (ms). As detailed in Table 3, the latency for non-iterative methods represents a single forward pass, while for iterative methods, we report the latency required to achieve optimal performance (T=60). When compared to similar diffusion-based models: Although our parameter count (47.16M) is comparable to that of Diffusion (45.37M), our SGDFuse (59ms) is significantly faster in terms of inference speed than both Diffusion and Text-Difuse. Furthermore, our practical latency (59ms) is also superior to many non-iterative methods, such as PIAFusion, and CDDFuse, and remains on the same order of magnitude as lightweight models like SAGE and TarDAL. In summary, this analysis demonstrates that our two-stage architecture (particularly the simplified diffusion task in Stage II) is computationally efficient. SGDFuse achieves a fast practical inference 
\begin{figure*}[htbp]
\centering
\includegraphics[scale=0.48]{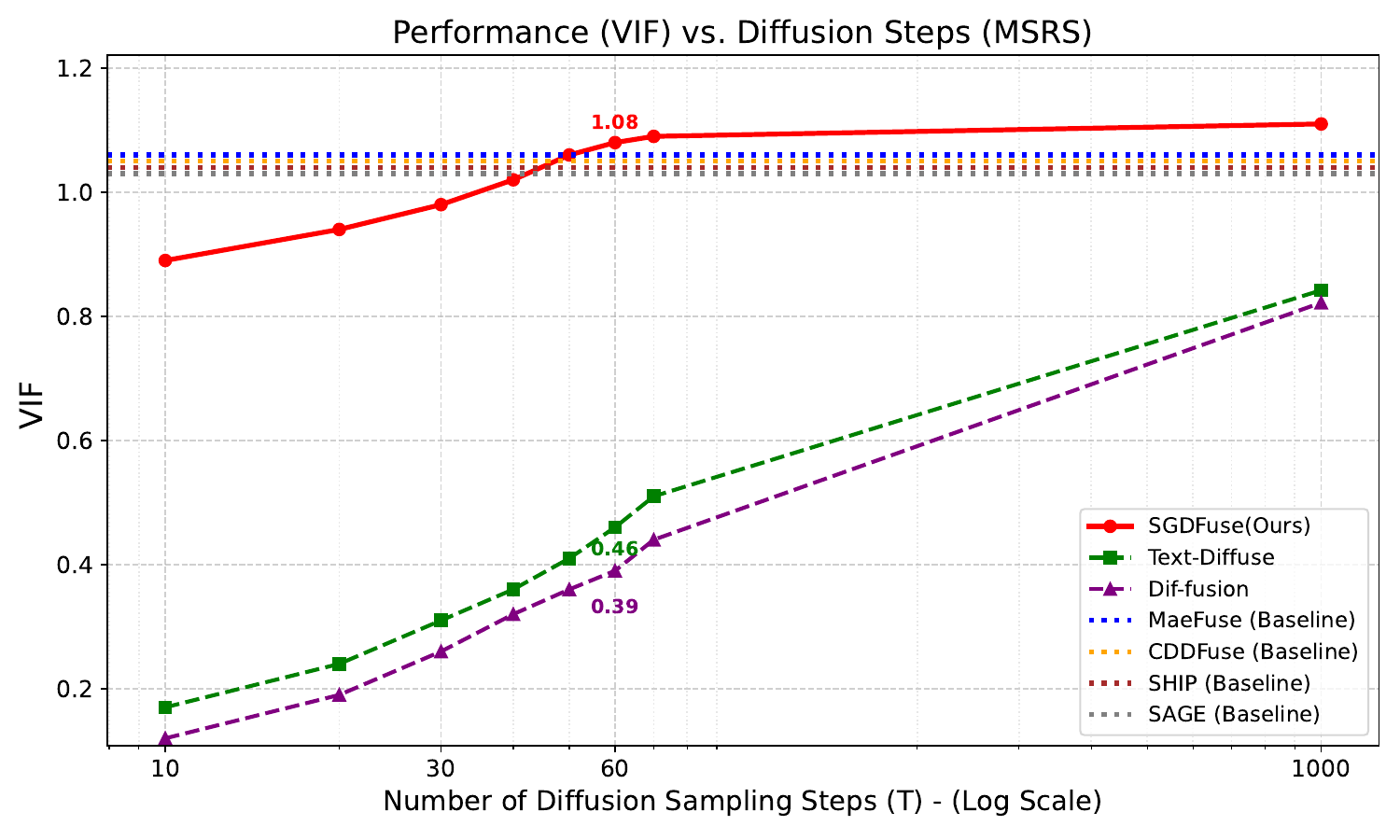}
\caption{Comparative Analysis of VIF Performance versus Diffusion Steps on the MSRS Dataset.}
\label{figure}
\end{figure*}
speed while maintaining a moderate computational cost (220.74G FLOPs). This proves that our method strikes an excellent trade-off between fusion performance and computational resource consumption, highlighting its strong competitiveness and practical applicability.
\begin{figure*}[htbp]
\centering
\includegraphics[scale=0.48]{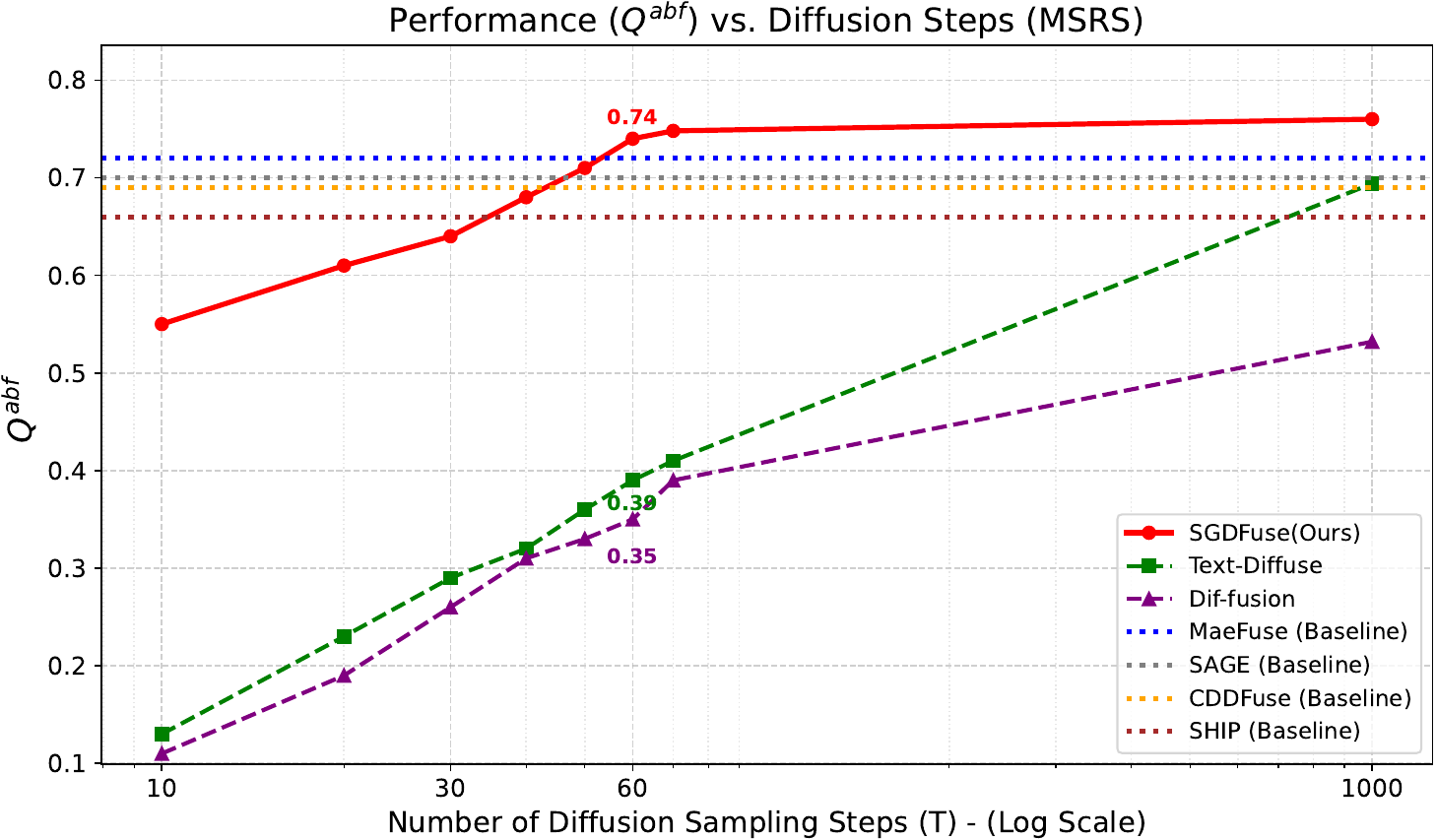}
\caption{Comparative Analysis of $Q^{abf}$ Performance versus Diffusion Steps on the MSRS Dataset.}
\label{figure}
\end{figure*}

\subsubsection{performance metrics vs. number of diffusion steps}  

To further analyze the relationship between the number of diffusion steps and fusion performance, we plotted the 'quality-versus-speed trade-off curve' for key performance metrics (VIF and $Q^{abf}$) as a function of sampling steps (T), and compared it against several representative SOTA methods, as shown in Figure 10 and Figure 11. Across all tested sampling steps, our method significantly and consistently outperforms other methods on both the VIF and $Q^{abf}$ metrics. Furthermore, it demonstrates excellent convergence efficiency. As illustrated in the figure, the model's performance rapidly reaches its peak at T=60 steps. At this relatively low step count, our performance already comprehensively surpasses the baseline levels of all other SOTA methods. This perfectly demonstrates the excellent balance our method achieves between fusion performance and practical efficiency.

\subsection{Ablation Study}

To thoroughly investigate the contribution of each component in the network architecture, a series of ablation studies were carefully designed and conducted, with systematic evaluations performed on the MSRS dataset. Furthermore, to validate the generalizability of our core design choices, we conduct an additional ablation study on the challenging low-light LLVIP dataset, which represents a significantly different scenario.

\subsubsection{Impact of SAM} The SAM offers high-quality semantic masks in stage II, guiding the diffusion model for structural refinement and semantic enhancement, thereby improving image integrity and consistency. To assess SAM’s effectiveness, we conducted ablation studies by removing or modifying its components, designing three variants to analyze the impact of semantic guidance on fusion performance.

(1). Variant (a): Removing the SAM module (w/o SAM)

In this setting, we completely remove the SAM module and replace the semantic masks with randomly cropped patches from the original images, thereby fully eliminating the guidance of semantic priors in the diffusion process. This allows us to evaluate the model’s performance in the absence of high-level semantic information.

(2). Variant (b): Removing the semantic mask of the infrared image (w/o IR SAM)

This variant retains only the semantic mask generated from the visible image while removing the semantic information of the infrared modality, aiming to analyze the specific role of thermal radiation structural features from the infrared image in guiding the optimization process.

(3). Variant (c): Removing the semantic mask of the visible image (w/o VIS SAM) Contrary to the previous setting, this variant retains the infrared semantic mask while removing the visible light mask, thereby evaluating the importance of the rich texture and edge information from the visible images in detail preservation and target recognition.
\begin{table}[!t]
\centering
\caption{Quantitative results of key ablation variants on the MSRS dataset. Bold numbers indicate the best performance.}
\setlength{\tabcolsep}{4.5pt} % 调整列间距
\renewcommand{\arraystretch}{1.39} % 控制行间距
\footnotesize 
\begin{tabular}{lccccccc}
\toprule
\textbf{MSRS} & \textbf{EN} & \textbf{SD} & \textbf{SF} & \textbf{MI} & \textbf{SCD} & \textbf{VIF} & \textbf{$Q^{abf}$} \\
\midrule

\rowcolor{lightgray}
\multicolumn{8}{c}{\textbf{I. Impact of SAM}} \\
w/o SAM & 6.43 & 43.27 & 11.98 & 2.54 & 1.48 & 0.87 & 0.62 \\
w/o IR SAM & 6.68 & 44.73 & 12.86 & 2.79 & 1.69 & 0.96 & 0.70 \\
w/o VIS SAM & 6.72 & 44.82 & 12.79 & 2.84 & 1.67 & 0.98 & 0.68 \\

\rowcolor{lightgray}
\multicolumn{8}{c}{\textbf{II. Impact of Two-Stage Training}} \\
w/o Stage I & 6.71 & 45.03 & 13.03 & 2.87 & 1.69 & 0.98 & 0.71 \\
w/o Stage II & 6.52 & 44.18 & 12.87 & 2.64 & 1.64 & 0.94 & 0.66 \\
End-to-End & 6.76 & 45.17 & 13.12 & 2.93 & 1.72 & 1.01 & 0.72 \\

\rowcolor{lightgray}
\multicolumn{8}{c}{\textbf{III. Impact of the Diffusion Process}} \\
w/o Dif & 6.69 & 44.89 & 12.96 & 2.83 & 1.62 & 0.99 & 0.63 \\

\rowcolor{lightgray}
\multicolumn{8}{c}{\textbf{IV. Impact of HFAH}} \\
w/o HFAF & 6.74 & 45.05 & 13.14 & 2.91 & 1.67 & 1.03 & 0.68 \\

\midrule
\rowcolor{lightblue}
\textbf{Ours method} & \textbf{6.81} & \textbf{45.28} & \textbf{13.27} & \textbf{2.99} & \textbf{1.73} & \textbf{1.08} & \textbf{0.74} \\
\bottomrule
\end{tabular}
\end{table}

As shown in Table 4, variant (a) without SAM performs worst in all metrics, highlighting the importance of semantic priors for structural and semantic fidelity. Variants (b) and (c) show milder drops but remain inferior to the full model, confirming the complementary benefits of dual-modality semantic masks. These results validate SAM's critical role in enhancing semantic representation and downstream task adaptability.
\begin{figure*}[htbp]
\centering
\includegraphics[scale=0.3]{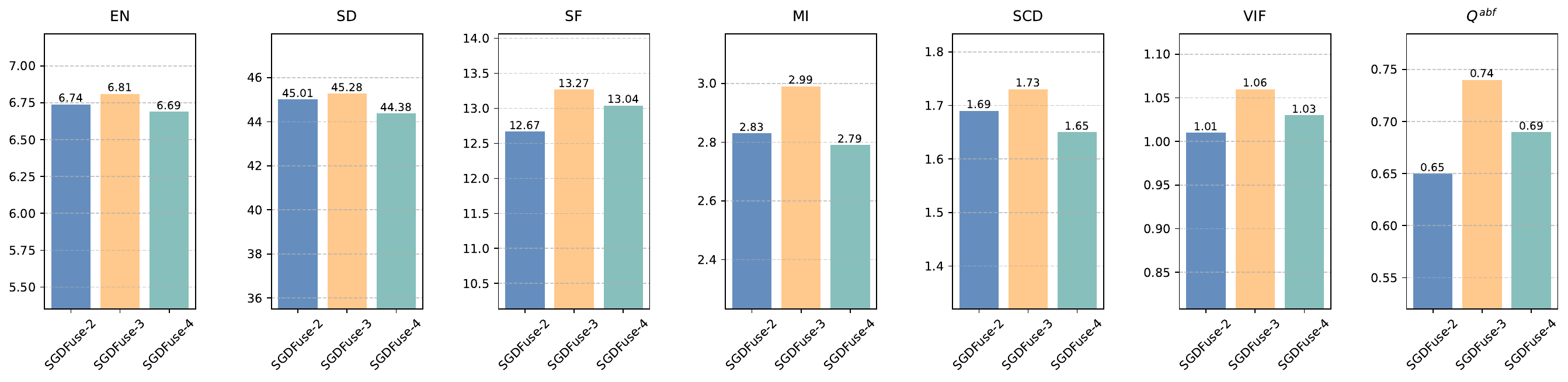}
\caption{Quantitative ablation results on the MSRS dataset regarding the number of repeated MSFEM and TB modules.}
\label{figure}
\end{figure*}

\subsubsection{Analysis of the Impact of the Number of MSFEM and TB Modules} To explore the optimal stacking depth for feature extraction, we performed ablation experiments by varying the number of MSFEM and TB modules in Stage I from 2 to 4 to investigate the influence of stacking depth on fusion performance. As shown in Figure 12, increasing the number of modules generally improves fusion quality by enhancing detail restoration and structural consistency. However, performance saturates or slightly declines beyond three layers due to potential redundancy and overfitting. Therefore, we adopt three MSFEM and three TB layers as the final configuration, striking a balance between accuracy and efficiency, and providing a strong feature foundation for stage II diffusion optimization.

\subsubsection{Impact of Two-Stage Training} The proposed SGDFuse adopts a deliberate two stage training strategy designed to decouple the fundamentally different objectives of structural alignment and generative refinement. To rigorously evaluate the scientific necessity of this design, we compared the full framework with several variants: a Stage I only model ($w/o$ Stage II), a Stage II only variant ($w/o$ Stage I), and a one stage End to End (E2E) architecture. In the E2E setting, the structural prior generation is bypassed, and a 6 channel tensor—concatenated from raw IR (1 channel), VIS (3  channel), and dual SAM masks (1 channel each)—is fed directly into the conditional diffusion network for training.
\begin{figure*}[htbp]
\centering
\includegraphics[scale=0.65]{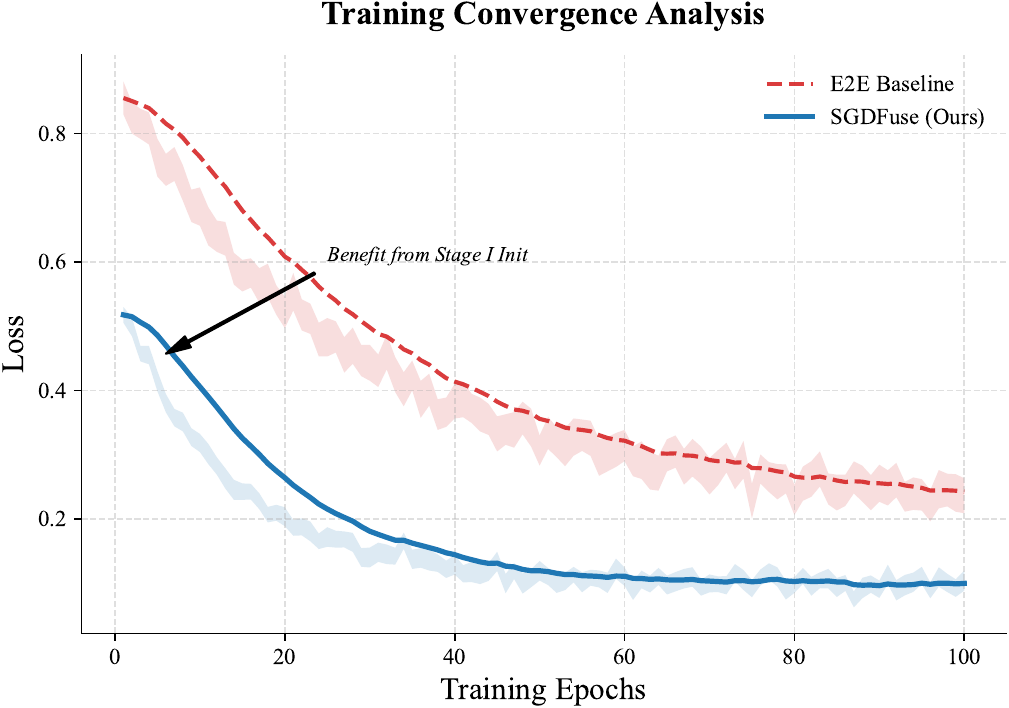}
\caption{Training convergence analysis of the proposed SGDFuse and the E2E baseline. The shaded areas represent the variance across multiple training sessions, the solid and dashed lines represent the smoothed mean values of the total loss.}
\label{figure}
\end{figure*}

As shown in Table 4 and Table 5, although the E2E variant utilizes all source information, its performance—particularly in $Q^{abf}$ and VIF—is consistently inferior to our two stage approach. This gap empirically validates a significant task conflict in E2E learning, where the network struggles to simultaneously optimize pixel-level cross-modal alignment and iterative semantic generation.

Further investigation of the training loss curves reveals that our two stage framework achieves faster convergence and a more stable trajectory than the E2E model. as shown in Figure 13. It is evident that our two stage approach achieves significantly faster convergence and maintains a more stable loss trajectory compared to the E2E model. This is because the E2E model must simultaneously navigate the complex landscape of cross-modal alignment and high-fidelity generation, leading to substantial gradient interference. In contrast, by providing a robust structural prior $(F_1)$ from Stage I, we effectively simplify the generative task for the diffusion model, allowing it to converge toward a more optimal solution with lower computational effort.

\subsubsection{Impact of the Diffusion Process} As the core of the second stage, the diffusion model refines structure and enhances semantics via progressive denoising. To verify its effect, we conducted an ablation by removing the diffusion process (w/o Dif) while keeping the network unchanged. Table 4 shows that on MSRS, all metrics decline without diffusion, demonstrating its critical role in improving clarity, structural integrity, and fusion quality under semantic guidance. This confirms diffusion modeling’s unique advantage and necessity for high-fidelity, semantically consistent multimodal fusion.

\subsubsection{Impact of HFAH} To enhance semantic guidance and structural restoration during denoising, we introduce the HFAH in the decoder of the diffusion model. HFAH aggregates multi-scale features via spatial attention to model edges and semantic regions more effectively. In ablation (w/o HFAH), removing hierarchical fusion and attention leads to noticeable drops in boundary fidelity and texture detail, as shown in Table 4—especially in complex scenes. These results confirm that HFAH improves fine-grained structural recovery and semantic consistency.
\begin{table}[!t]
\centering
\caption{Quantitative results of key ablation variants on the LLVIP dataset. Bold numbers indicate the best performance.}
\setlength{\tabcolsep}{4.5pt} % 调整列间距
\renewcommand{\arraystretch}{1.39} % 控制行间距
\footnotesize 
\begin{tabular}{lccccccc}
\toprule
\textbf{LLVIP} & \textbf{EN} & \textbf{SD} & \textbf{SF} & \textbf{MI} & \textbf{SCD} & \textbf{VIF} & \textbf{$Q^{abf}$} \\
\midrule

\rowcolor{lightgray}
\multicolumn{8}{c}{\textbf{I. Impact of SAM}} \\
w/o SAM & 6.87 & 42.26 & 15.69 & 2.52 & 1.43 & 0.83 & 0.63 \\
w/o IR SAM & 6.95 & 42.75 & 15.87 & 2.68 & 1.60 & 0.85 & 0.64 \\
w/o VIS SAM & 6.98 & 43.01 & 15.96 & 2.73 & 1.56 & 0.87 & 0.66 \\

\rowcolor{lightgray}
\multicolumn{8}{c}{\textbf{II. Impact of Two-Stage Training}} \\
w/o Stage I & 6.91 & 42.94 & 15.87 & 2.60 & 1.55 & 0.93 & 0.64 \\
w/o Stage II & 6.82 & 42.79 & 15.39 & 2.59 & 1.53 & 0.91 & 0.62 \\
End-to-End & 7.01 & 43.14 & 16.03 & 2.68 & 1.61 & 0.94 & 0.65 \\

\rowcolor{lightgray}
\multicolumn{8}{c}{\textbf{III. Impact of the Diffusion Process}} \\
w/o Dif & 6.99 & 43.07 & 15.98 & 2.65 & 1.54 & 0.92 & 0.65 \\

\rowcolor{lightgray}
\multicolumn{8}{c}{\textbf{IV. Impact of HFAH}} \\
w/o HFAF & 7.06 & 43.19 & 16.11 & 2.72 & 1.61 & 0.96 & 0.69 \\

\midrule
\rowcolor{lightblue}
\textbf{Ours method} & \textbf{7.12} & \textbf{43.35} & \textbf{16.15} & \textbf{2.79} & \textbf{1.65} & \textbf{0.99} & \textbf{0.71} \\
\bottomrule
\end{tabular}
\end{table}

\subsubsection{Generalizability Validation on LLVIP Dataset} To further validate the robustness and generalizability of our core design choices, we conducted an additional ablation study on the LLVIP dataset. This dataset, containing primarily nighttime scenes with weak textures and extremely low illumination , provides a significantly different and challenging scenario compared to the MSRS dataset. We re-evaluated our most critical components, with the results presented in Table 5. The results strongly mirror our findings from Table 4. The full model (Ours method) achieves the best performance. Notably, the w/o Stage II variant shows a severe degradation in all metrics, confirming that our diffusion-based refinement stage is essential for high-fidelity reconstruction across scenarios. Similarly, the w/o SAM and w/o HFAH variants also perform significantly worse than the full model, demonstrating that semantic guidance and hierarchical feature aggregation are robust and critical contributions. This consistent performance across both MSRS and LLVIP datasets confirms that the contributions of our core components are robust and generalizable.

\subsubsection{Analysis of SAM’s Applicability in the Infrared Domain} To address the concern regarding SAM’s cross-modal generalization, we provide a systematic analysis combining qualitative visualization (Figure 2) and quantitative ablation results (Table 4 and Table 5). While SAM is primarily pre-trained on visible datasets, its powerful zero-shot generalization capability allows it to identify thermal targets in the IR domain. As qualitatively demonstrated in Figure 2, SAM produces high-precision semantic masks that accurately delineate thermal boundaries, even in extreme low-light scenarios where visible information is severely degraded.

Quantitative results further support this. As shown in Table 5 (LLVIP), which features challenging nighttime scenes, the removal of IR SAM masks ($w/o$ IR SAM) leads to a substantial drop in $Q^{abf}$ from 0.71 to 0.64. Notably, this degradation is more pronounced than that observed when removing VIS SAM masks (where $Q^{abf}$ drops to 0.66). A similar trend is observed on the MSRS dataset (Table 4), where $Q^{abf}$ falls from 0.74 to 0.70 without IR SAM. These results empirically prove that IR SAM masks provide critical structural priors that VIS SAM cannot compensate for, especially in low-light environments.

The effectiveness of SAM in the IR domain stems from its architectural sensitivity to sharp intensity gradients and closed contours. In IR imagery, thermal targets typically exhibit high contrast against the background. SAM effectively leverages these strong thermal gradients to generate semantically consistent masks. These masks serve as 
\begin{figure*}[htbp]
\centering
\includegraphics[scale=0.9]{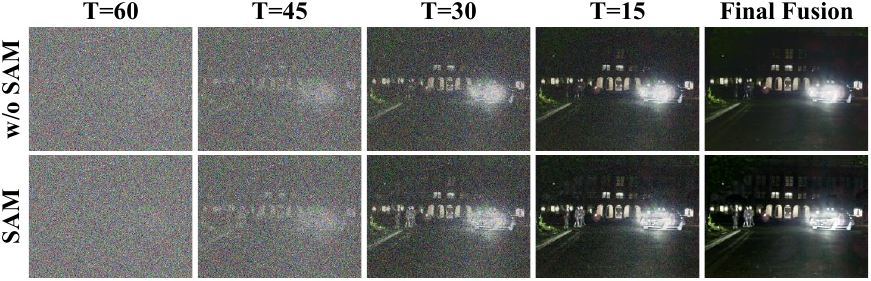}
\caption{Visualization comparison of the diffusion denoising process with (SAM) and without (w/o SAM) SAM guidance.}
\label{figure}
\end{figure*}
essential "spatial anchors" for the Stage II diffusion process, ensuring the preservation of critical target structures during generative refinement. This analysis confirms that the dual-modal SAM integration is a scientifically grounded strategy rather than a mere engineering heuristic.

\subsubsection{Semantically Guided Analysis of Diffusion Processes} To deeply analyze how SAM-based guidance influences the fusion process during diffusion, we visualized intermediate steps in the reconstruction of diffusion models. We compared the full model (SAM) with its variant without SAM. As shown in Figure 14, we observe that in the early stages ($T=60$, $T=45$), both trajectories are (as expected) dominated by noise. However, a significant divergence emerges at $T=30$: the full SAM model begins constructing coherent semantic structures for the left figure and distinct contours for the right vehicle, while the non-SAM model remains blurred noise at identical strides, failing entirely to capture these semantic targets. This semantic advantage persists through $T=15$ and into the final fusion stage. This visualization clearly demonstrates that our SAM-guided framework effectively steers the denoising trajectory toward semantically correct solutions from early stages, thereby revealing the model's internal mechanisms with greater clarity.
\begin{table}[htbp]
\centering
\caption{Quantitative robustness analysis on the M$^3$FD dataset using perturbed SAM masks. Bold numbers indicate the best performance.}
\setlength{\tabcolsep}{4.5pt} % 调整列间距
\renewcommand{\arraystretch}{1.39} % 控制行间距
\footnotesize 
\begin{tabular}{lccccccc}
\toprule
\textbf{Perturbation} & \textbf{EN} & \textbf{SD} & \textbf{SF} & \textbf{MI} & \textbf{SCD} & \textbf{VIF} & \textbf{$Q^{abf}$} \\
\midrule
Eroded Mask & 6.93 & 37.91 & 13.19 & 2.93 & 1.64 & 0.97 & 0.65 \\
Dilated Mask & 6.89 & 37.85 & 13.14 & 2.91 & 1.61 & 0.95 & 0.62 \\
Original mask & \textbf{6.99} & \textbf{38.02} & \textbf{13.27} & \textbf{2.99} & \textbf{1.69} & \textbf{1.01} & \textbf{0.69} \\
\bottomrule
\end{tabular}
\end{table}

\subsection{Robustness and Generalizability Analysis}

To comprehensively assess the stability and applicability of the proposed framework, this section conducts an in-depth analysis of two critical aspects: (1) the framework's robustness to inaccuracies in semantic priors; and (2) its generalizability to different sources of semantic priors.

\subsubsection{Robustness Analysis to SAM Prior Inaccuracies} To address the concern that our framework's performance may hinge on the accuracy of SAM-generated masks, we conducted a robustness experiment on the MSRS dataset. We simulated two common types of segmentation inaccuracies mentioned by the reviewer: "incomplete masks" (False Negatives) and "mis-segmented masks" (False Positives). To simulate incomplete masks and mis-segmented masks, we applied Morphological Erosion with a $7 \times 7$ kernel to the original SAM masks ($M_{ir}, M_{vis}$). We then fed these "perturbed" masks into our frozen, pre-trained Stage II model to evaluate the impact on fusion quality.

This robustness analysis is visualized in Figure 15
\begin{figure*}[htbp]
\centering
\includegraphics[scale=0.85]{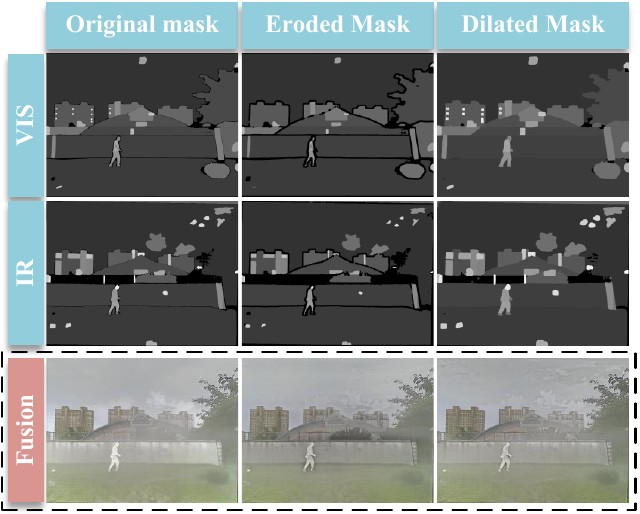}
\caption{Qualitative analysis of robustness to perturbed SAM masks.}
\label{figure}
\end{figure*}
and quantitatively evaluated in Table 6. Figure 15 clearly demonstrates that the $7 \times 7$ kernel causes significant perturbations along the mask boundary, yet the corresponding fusion results still exhibit high robustness. Despite minor visual changes compared to the original results, the model did not experience catastrophic failure nor produce the expected “false fusion emphasis” phenomenon. Key targets (e.g., figures) are preserved, and background textures remain stable. The quantitative analysis in Table 6 corroborates this conclusion. As anticipated, all metrics exhibit predictable declines after introducing the noise mask, confirming that semantic priors indeed guide the fusion process. The key finding, however, is that performance degradation is gradual rather than abrupt. For instance, the $Q^{abf}$ metric measuring structural fidelity decreased only from 0.69 to 0.65 (erosion) and 0.62 (dilatation). Similarly, the visual information fidelity (VIF) metric decreased only from 1.01 to 0.97 and 0.95. This experiment demonstrates through both visual and quantitative validation that while our framework significantly benefits from precise prior information, it is not overly fragile and maintains high robustness in the face of segmentation errors.
\begin{table}[!t]
\centering
\caption{Generalizability Analysis of the Framework to Different Semantic Priors on the MSRS Dataset. Bold numbers indicate the best performance.}
\setlength{\tabcolsep}{4.5pt} % 调整列间距
\renewcommand{\arraystretch}{1.39} % 控制行间距
\footnotesize 
\begin{tabular}{lccccccc}
\toprule
\textbf{Method} & \textbf{EN} & \textbf{SD} & \textbf{SF} & \textbf{MI} & \textbf{SCD} & \textbf{VIF} & \textbf{$Q^{abf}$} \\
\midrule
w/o SAM & 6.43 & 43.27 & 11.98 & 2.54 & 1.48 & 0.87 & 0.62 \\
w/ Mask2Former & 6.72 & 44.95 & 12.97 & 2.90 & 1.67 & 1.03 & 0.71 \\
w/ SegFormer & 6.70 & 44.83 & 12.88 & 2.87 & 1.64 & 1.01 & 0.70 \\
w/ SAM & \textbf{6.81} & \textbf{45.28} & \textbf{13.27} & \textbf{2.99} & \textbf{1.73} & \textbf{1.08} & \textbf{0.74} \\
\bottomrule
\end{tabular}
\end{table}

\subsubsection{Generalizability Analysis to Alternative Semantic Priors} To verify whether our “semantic guidance” concept is inherently generalizable or merely fragilely dependent on a specific tool (i.e., SAM), we conducted a “substitute prior” experiment. Specifically, we replaced SAM with Mask2Former and SegFormer, both pretrained on large-scale datasets such as COCO, as alternative sources of semantic priors. All other components of our model (including pretrained weights) were kept unchanged, and we re-evaluated the framework on the MSRS 
\begin{figure*}[htbp]
\centering
\includegraphics[scale=0.75]{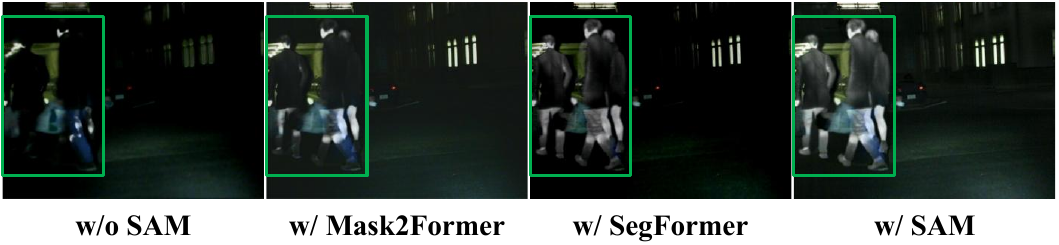}
\caption{Visual comparison of fusion results using different semantic guidance (Mask2Former, SegFormer, SAM) versus the unguided baseline (w/o SAM). The green boxes highlight the regions of interest (pedestrian targets) for clearer comparison of semantic preservation.}
\label{fig15}
\end{figure*}
dataset. As shown in Table 7, using Mask2Former \cite{39} or SegFormer \cite{40} as guiding priors yielded significant performance improvements compared to the “no-guidance” baseline (w/o SAM). To visually demonstrate this point, we also conducted a qualitative comparison, as clearly shown in Figure 16. The 'w/o SAM' (unguided) model fails to preserve critical target structures (such as pedestrians), whereas all three 'guided' models (w/ Mask2Former, w/ SegFormer, and w/ SAM) successfully and robustly highlight the targets. This strongly demonstrates that the core idea of semantic guidance is indeed general and effective, rather than being narrowly tied to SAM. Moreover, the full version with SAM achieved the best overall performance across all key metrics. This indicates that, owing to its powerful open-set generalization capability, SAM provides the highest-quality semantic priors for this fusion task, thereby validating the rationale of adopting it as our preferred semantic guidance tool.

\subsection{Visual Analysis with Intensity Profiles} To provide a granular quantitative visualization of fusion quality, we select a representative nighttime scene from the LLVIP dataset for intensity profile analysis. As illustrated in Figure 17, we plot the pixel intensity across salient pedestrians (indicated by the red solid line) to compare our method with several state-of-the-art competitors. The profile reveals that SGDFuse achieves a superior synergy between thermal radiation preservation and edge fidelity: our method maintains intensity peaks that align closely with the infrared source while producing significantly steeper gradients at target boundaries. This pixel-level evidence validates that the SGDFuse framework effectively suppresses common pitfalls such as background over-enhancement and thermal suppression, ensuring high-fidelity reconstruction for complex real-world scenarios.
\begin{figure*}[htbp]
\centering
\includegraphics[scale=0.54]{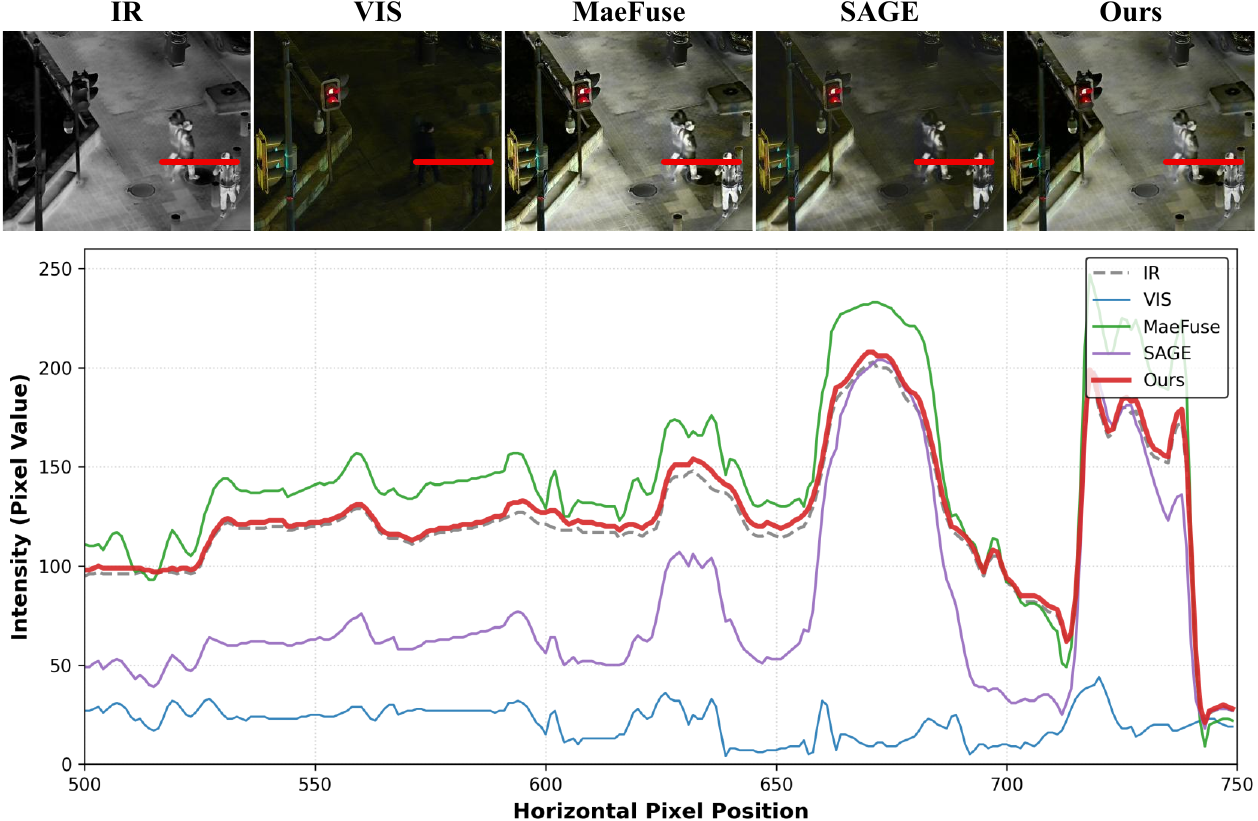}
\caption{Intensity profile analysis on a representative scene from the LLVIP dataset. The red solid lines in the top row indicate the sampling positions.}
\label{figure}
\end{figure*}
\begin{table*}[htbp]
\centering
\caption{Quantitative comparison on MRI-PET and MRI-SPECT datasets, with red representing the best results, blue representing the second best results.}
\resizebox{1\textwidth}{!}{
\begin{tabular}{l|ccccccc|ccccccc}
\toprule
\multirow{2}{*}{\textbf{Method}} & \multicolumn{7}{c|}{\textbf{MRI-PET}} & \multicolumn{7}{c}{\textbf{MRI-SPECT}} \\
\cmidrule(lr){2-8} \cmidrule(lr){9-15}
& \textbf{EN} & \textbf{SD} & \textbf{SF} & \textbf{MI} & \textbf{SCD} & \textbf{VIF} & \textbf{$Q^{abf}$}& \textbf{EN} & \textbf{SD} & \textbf{SF} & \textbf{MI} & \textbf{SCD} & \textbf{VIF} & \textbf{$Q^{abf}$}\\
\midrule
\textbf{U2Fusion}\cite{1} & 3.48 & 55.46 & 22.17 & 1.61 & 1.22 & 0.35 & 0.47 & 3.44 & 54.24 & 18.67 & 1.66 & 1.18 & 0.51 & 0.59 \\
\textbf{CDDFuse} \cite{37} & 4.37 & 77.28 & 28.07 & 1.82 & 1.79 & 0.60 & 0.64 & 4.02 & 63.45 & 20.28 & 1.93 & 1.69 & 0.67 & 0.68 \\\
\textbf{Dif-fusion}\cite{5} & 4.34 & 74.12 & 27.13 & 1.81 & 1.66 & 0.59 & 0.64 & 3.95 & 63.31 & 20.10 & 1.83 & 1.64 & 0.63 & 0.67\\
\textbf{SHIP} \cite{24} & 5.34 & 67.96 & 26.33 & 1.84 & 1.53 & 0.60 & 0.65 & 4.79 & 52.20 & 19.77 & 1.88 & 0.67 & 0.60 & 0.68 \\
\textbf{Text-Difuse}\cite{11} & 5.07 & 76.49 & 25.74 & 1.73 & 1.71 & 0.46 & 0.34 & 4.81 & 64.20 & 17.10 & 1.62 & 1.60 & 0.44 & 0.26 \\
\textbf{MaeFuse} \cite{6} & 5.19 & {\color[HTML]{0000FF}77.39} & {\color[HTML]{FF0000} 28.41} & 1.83 & 1.75 & {\color[HTML]{0000FF}0.62} & {\color[HTML]{0000FF}0.69} & {\color[HTML]{0000FF}4.85} & 64.56 & {\color[HTML]{FF0000}20.33} & 1.92 & 1.71 & 0.66 & 0.68\\
\textbf{SAGE }\cite{12} & {\color[HTML]{FF0000}5.32} & 77.31 & 28.12 & {\color[HTML]{0000FF}1.85} & {\color[HTML]{0000FF}1.76} & 0.61 & 0.67 & {\color[HTML]{FF0000}4.87} & {\color[HTML]{FF0000}64.79} & 20.16 & {\color[HTML]{0000FF}1.95} & {\color[HTML]{FF0000}1.74} & {\color[HTML]{0000FF}0.69} & {\color[HTML]{0000FF}0.69}\\
\midrule
\textbf{SGDFuse(Ours)} & {\color[HTML]{0000FF} 5.24} & {\color[HTML]{FF0000} 77.61} & {\color[HTML]{0000FF} 28.19} & {\color[HTML]{FF0000} 1.89} & {\color[HTML]{FF0000} 1.77} & {\color[HTML]{FF0000} 0.64} &{\color[HTML]{FF0000} 0.71} & 4.82 & {\color[HTML]{0000FF}64.62} & {\color[HTML]{0000FF}20.31} & {\color[HTML]{FF0000}1.98} & {\color[HTML]{0000FF} 1.73 }& {\color[HTML]{FF0000}0.72} & {\color[HTML]{FF0000}0.70} \\
\bottomrule
\end{tabular}}
\end{table*}

\subsection{Validation on Medical Multimodal Fusion Domains} To validate the robustness and generalization capabilities of our proposed SGDFuse framework beyond the IVIF domain, we conducted 
\begin{figure*}[htbp]
\centering
\includegraphics[scale=0.575]{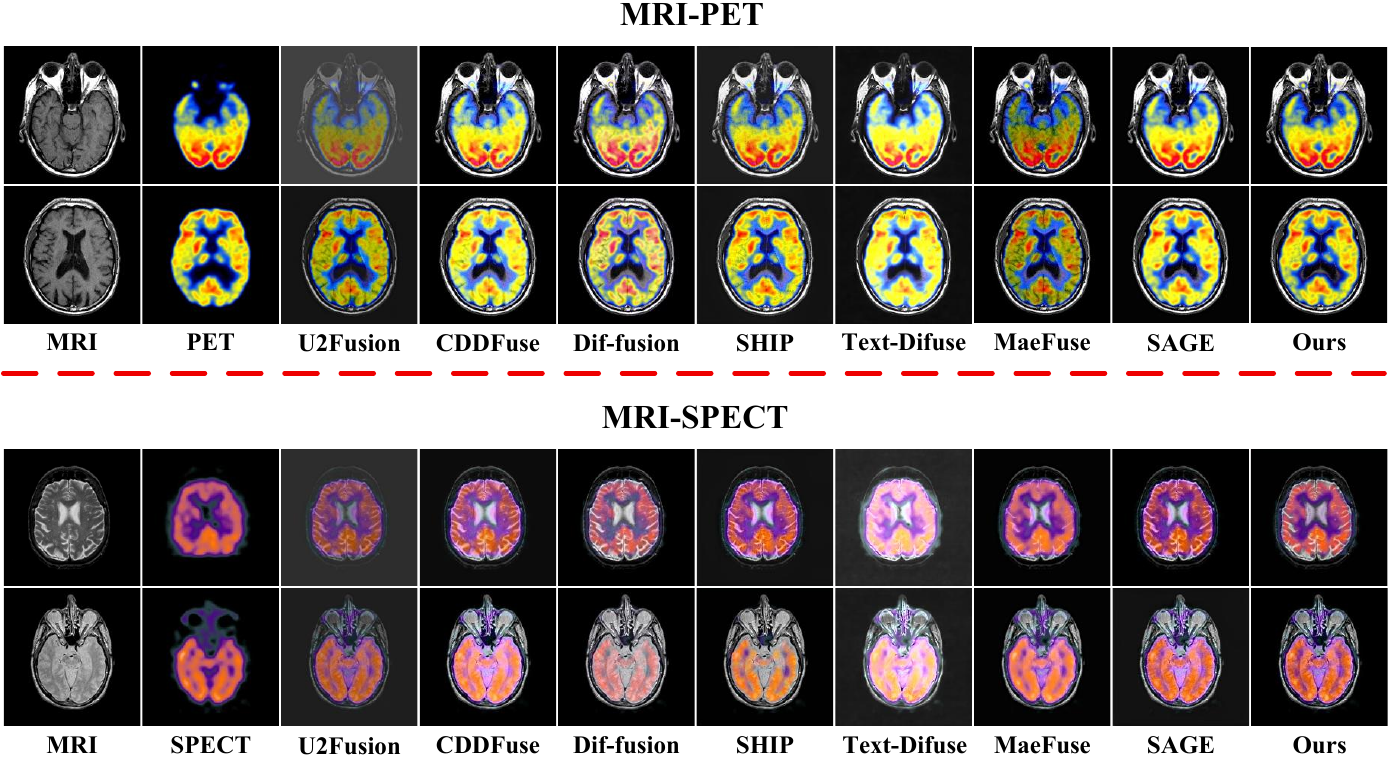}
\caption{Visual comparison of fusion results by our method against other representative state-of-the-art methods on the MRI-PET and MRI-SPECT datasets.}
\label{figure}
\end{figure*}
additional cross-domain experiments on two public medical image fusion datasets: MRI-PET and MRI-SPECT. Using the standard training-test split from the Harvard Medical website, all baseline methods were retrained from scratch under identical experimental settings to ensure a rigorous and fair comparison. This ensures that all models were specifically optimized for the unique characteristics of medical imaging modalities, rather than relying on out-of-domain IVIF pretrained weights. Quantitative results are presented in Table 8, while qualitative comparisons are shown in Figure 18. The results demonstrate our framework's robust generalization capability. On the MRI-PET dataset, SGDFuse achieves state-of-the-art performance on key metrics including MI, VIF, and $Q^{abf}$. On the MRI-SPECT dataset, our model remains highly competitive. Figure 18 demonstrates that our original fusion method preserves finer structures and details in both PET and MRI images, showcasing its superior capability in retaining intricate details and enhancing overall image quality. This experiment confirms that our core SAM-guided generation Methodological Framework is not confined to IVIF solutions but represents a robust and generalizable framework that effectively adapts to other challenging domains, such as medical image fusion.
\begin{figure*}[t]
\centering
\includegraphics[scale=0.31]{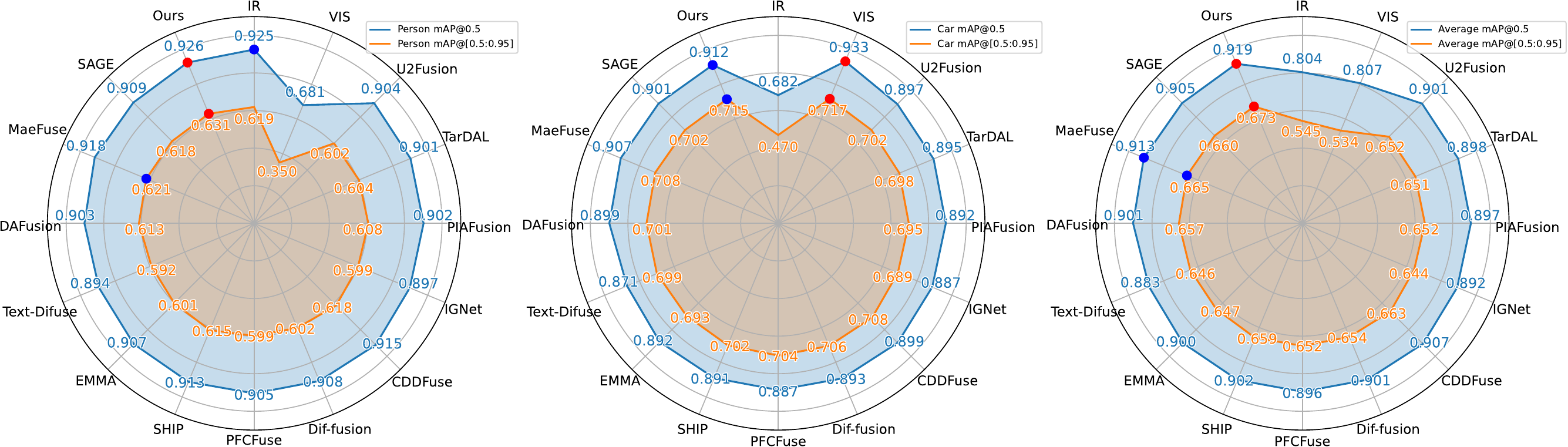}
\caption{Quantitative evaluation of fusion methods on object detection. Red and blue dots indicate the best and second-best results, respectively.}
\label{figure}
\end{figure*}
\begin{figure*}[htbp]
\centering
\includegraphics[scale=0.52]{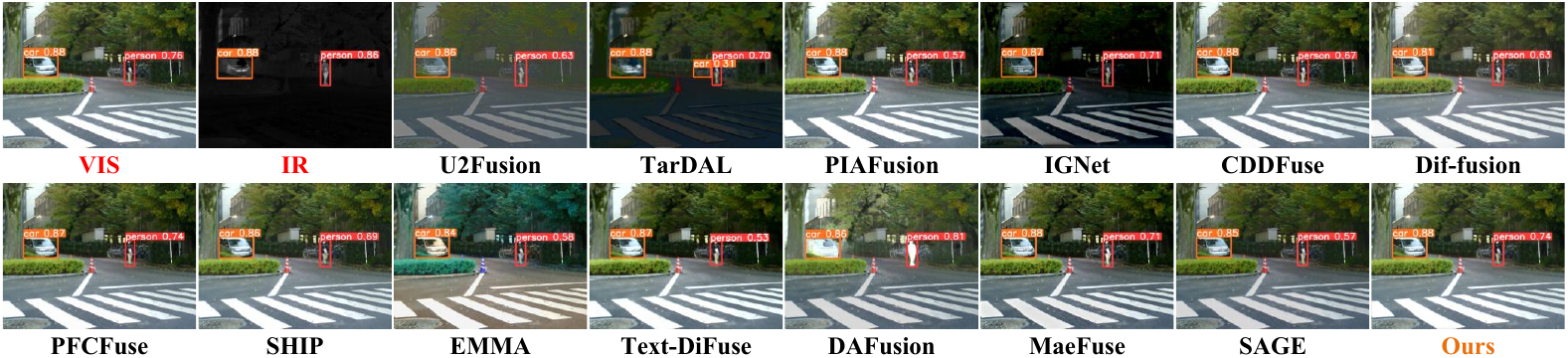}
\caption{Visual results of the object detection task on the MSRS dataset.}
\label{figure}
\end{figure*}

\subsection{Extension to high-level vision task}

To validate adaptability in high-level vision tasks, we applied the method to semantic segmentation and object detection, using their performance to comprehensively evaluate fusion quality and demonstrate task-driven advantages.

\subsubsection{Object Detection} Object detection experiments on the MSRS dataset using YOLOv5 \cite{41} show that fused images outperform single-modality inputs. Our method achieves the best detection accuracy among mainstream fusion approaches, demonstrating its superior effectiveness and practical value in downstream tasks.

Figure 19 shows that fused images significantly improve detection accuracy over single-modality inputs. Our method outperforms mainstream fusion approaches in both “Person” and “Car” categories, demonstrating superior object perception and generalization. As shown in 
\begin{table*}[ht]
\centering
\caption{Quantitative comparison of different fusion methods across categories. Best and second-best scores are marked in \textcolor{red}{red} and \textcolor{blue}{blue}, respectively.}
\resizebox{0.9\textwidth}{!}{%
\large
\begin{tabular}{lcccccccccccccc}
\toprule
\textbf{Category} & Background & Car & Person & Curve & Color Cone & Bike & \textbf{mIoU}\\
\midrule
U2Fusion & 0.976 & 0.857 & 0.682 & 0.703 & 0.725 & 0.716 & 0.776 \\
TarDAL & 0.969 & 0.867 & 0.681 & 0.755 & 0.783 & 0.769 & 0.804 \\
PIAFusion & 0.972 & 0.871 & 0.680 & 0.721 & 0.766 & 0.739 & 0.791 \\
IGNet & 0.971 & 0.869 & 0.679 & 0.769 & 0.779 & 0.754 & 0.804 \\
CDDFuse & 0.975 & 0.872 & 0.685 & \textcolor{red}{0.789} & 0.794 & 0.785 & \textcolor{blue}{0.817} \\
Dif-fusion & 0.976 & 0.871 & 0.684 & 0.774 & 0.772 & 0.786 & 0.810 \\
PFCFuse & 0.975 & 0.873 & 0.683 & 0.737 & 0.748 & 0.729 & 0.791 \\
SHIP & 0.976 & 0.871 & 0.681 & 0.745 & 0.760 & 0.742 & 0.796 \\
EMMA & 0.977 & 0.862 & 0.656 & 0.735 & 0.769 & 0.638 & 0.773 \\
Text-Difuse & 0.972 & 0.873 & 0.673 & 0.743 & 0.764 & 0.748 & 0.796 \\
DAFusion & 0.977 & 0.868 & 0.679 & 0.756 & 0.778 & 0.750 & 0.802 \\
MaeFuse & 0.975 & 0.871 & \textcolor{blue}{0.685} & 0.774 & \textcolor{blue}{0.796} & \textcolor{blue}{0.787} & 0.814 \\
SAGE & \textcolor{red}{0.981} & \textcolor{blue}{0.879} & 0.675 & 0.747 & 0.783 & 0.774 & 0.806 \\
\midrule
Ours & \textcolor{blue}{0.979} & \textcolor{red}{0.879} & \textcolor{red}{0.686} & \textcolor{blue}{0.780} & \textcolor{red}{0.798} & \textcolor{red}{0.789} & \textcolor{red}{0.819} \\
\bottomrule
\end{tabular}%
}
\label{tab:fusion_comparison}
\end{table*}
Figure 20, our method outperforms others in accurately localizing multiple targets by effectively preserving infrared thermal information and integrating visible details, thereby enhancing detection accuracy, robustness, and adaptability in complex scenes.

\subsubsection{Semantic Segmentation} We use DeeplabV3+ \cite{42} for semantic segmentation evaluation, measuring performance with pixel-level IoU. DeeplabV3+ is a well-established model with strong feature modeling, making it suitable to assess how well the fused images support downstream segmentation tasks.

Table 9 shows segmentation results for six categories (Car, Person, Bike, Color Cone, Background, Curve). Our method achieves the highest IoU in most classes—especially Background, Car, and Person—and significantly outperforms others in mean IoU, demonstrating strong generalization and semantic consistency across categories.

Furthermore, Figure 21 shows semantic segmentation visualizations on complex scenes. U2Fusion and CDDFuse exhibit blurred boundaries and texture loss, struggling with multiple targets and fine details. MaeFuse and SAGE improve coherence but still missegment small objects and complex backgrounds. Our method achieves clearer boundaries 
\begin{figure*}[htbp]
\centering
\includegraphics[scale=0.275]{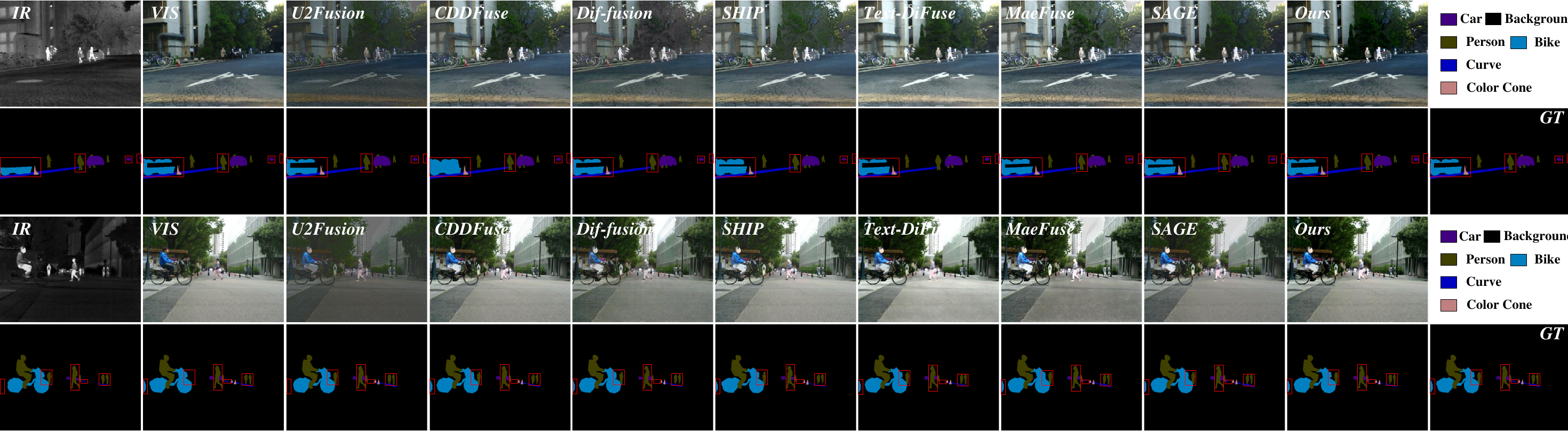}
\caption{Visualization results of semantic segmentation. We compare our method with seven classical approaches for qualitative analysis. The regions with significant differences are highlighted in red boxes.}
\label{figure}
\end{figure*}
and more complete contours, demonstrating superior structural fidelity and semantic consistency.

\section{DISCUSSION}
\noindent 

This paper proposes SGDFuse, an innovative framework for infrared and visible image fusion, designed to address the prevalent semantic gap in current fusion methods. Our core contribution lies in establishing a novel methodological framework: leveraging prior knowledge from SAM as explicit semantic guidance to constrain and optimize the generative process of a conditional diffusion model, thereby achieving high-fidelity synthesis of cross-modal information. Extensive quantitative experiments, along with evaluations on downstream tasks such as object detection and semantic segmentation, consistently demonstrate that the fused images generated by SGDFuse significantly surpass state-of-the-art methods in terms of information integrity, detail fidelity, and semantic consistency. This robustly validates the superiority and effectiveness of our proposed SGG  framework.

We have carefully considered practical implementation aspects, particularly the trade-off between performance and efficiency. The introduction of SAM and a diffusion model in SGDFuse admittedly increases computational complexity. However, we argue that this design is a necessary approach to overcome the fundamental limitations of traditional methods. Because traditional methods cannot accurately distinguish between foreground targets and background textures, they often lead to the loss of critical thermal radiation information or detail distortion, which fundamentally compromises the quality of the fused information. In contrast, SGDFuse exchanges a manageable computational overhead for a substantial leap in the semantic fidelity of the fused information. particularly the trade-off between performance and efficiency. In terms of real-time applicability, SGDFuse exchanges a manageable computational overhead for a substantial leap in semantic fidelity. With an optimized inference latency of 59ms, the framework demonstrates strong potential for deployment in high-stakes practical applications such as autonomous driving and intelligent surveillance.

Despite SGDFuse achieving SOTA performance, there remain broad avenues for future exploration. While SGDFuse serves as a specific and effective implementation, the underlying SGG logic should be viewed as a broader methodological approach for multi-modal fusion. The generalizability of this framework is evidenced by its compatibility with alternative semantic priors (e.g., Mask2Former and SegFormer in Section 4.5.2) and its successful application in the medical imaging domain (Section 4.7). Therefore, our current "formally simple yet effective" implementation should be regarded as one effective path, or a powerful baseline, within the SGG methodological framework.

Future exploration can proceed along several main directions: First, one could explore more 'complex' or 'sophisticated' guidance mechanisms. For instance, by designing a lightweight 'semantic encoder' to replace simple concatenation, thereby encoding semantic information into more refined features; or by injecting multi-scale semantic guidance at each timestep of the diffusion process. Second, regarding guidance quality, given that SAM's pre-training data is primarily based on visible light images, its generalization capability to the infrared domain—though proven effective in our experiments—still has room for improvement. In the future, the precision of the semantic guidance could be further enhanced through domain-adaptive fine-tuning. Third, future work will focus on model lightweighting, for instance, through knowledge distillation or by designing efficient samplers, to develop a version suitable for real-time applications and thereby broaden its operational boundaries.

\section{CONCLUSION}
\noindent 

Existing IVIF methods generally face a bottleneck in deeply understanding complex scene semantics. To address this, we propose SGDFuse, a novel framework that reframes IVIF as a SGG methodological task. By strategically decoupling structural alignment from generative refinement through a synergistic two-stage design, our method effectively resolves the inherent task conflict between cross-modal feature extraction and high-fidelity reconstruction. Guided by explicit SAM-based semantic priors and a tailored Mask-Guided Loss, SGDFuse enables the simultaneous preservation of thermal target saliency and fine-grained textural details. Extensive evaluations across multiple benchmarks and downstream tasks—including object detection and semantic segmentation—consistently demonstrate our method's superiority in both visual quality and task-level performance. Furthermore, the robust generalization of the SGG framework is validated through successful cross-domain applications in medical imaging. Future work will focus on optimizing the framework for real-time inference and exploring more sophisticated semantic-aware guidance mechanisms.

\section*{Acknowledgments}
This work is supported by the National Natural Science Foundation of China under grant Nos. U24A20219, 62272281, 62576193, 62202268, and 62301105; special funds of Taishan Scholars Project of Shandong Province under grant Nos. tsqn202507240; Natural Science Foundation of Shandong Province under grant Nos. ZR2025QC712, ZR2025QC695 and ZR2025MS985.

\section*{Data Availability Statement}
Data available on request from the authors.

\section*{Declaration of competing interest}
The authors declare that they have no known competing financial interests or personal relationships that could have appeared to influence the work reported in this paper.

\bibliographystyle{unsrt}
\bibliography{ref}

\end{sloppypar}
\end{document}